\documentclass[journal]{IEEEtran}

\ifCLASSINFOpdf
\else
\fi

\usepackage{times}
\usepackage{epsfig}
\usepackage{graphicx}
\usepackage{amsmath}
\usepackage{amssymb}

\usepackage{tabu}
\usepackage{multirow}
\usepackage{makecell}

\usepackage[switch]{lineno} % line number

\usepackage[pagebackref=true,breaklinks=true,colorlinks,bookmarks=false]{hyperref}

\begin{document}
% \linenumbers
\title{Perception-and-Regulation Network \\ for Salient Object Detection}

\author{Jinchao Zhu,
        Xiaoyu Zhang,
        Xian Fang,
        Junnan Liu % <-this % stops a space
\thanks{J. Zhu and X. Zhang are with College of Artificial Intelligence, Nankai University. X. Zhang is corresponding author.}
\thanks{X. Fang is with College of Computer Science, Nankai University.}
\thanks{Junnan Liu is with College of Intelligent Systems Science and Engineering, Harbin Engineering University.}}

%M. Shell was with the Department
%of Electrical and Computer Engineering, Georgia Institute of Technology, Atlanta,

\markboth{}%
{Shell \MakeLowercase{\textit{et al.}}: Bare Demo of IEEEtran.cls for IEEE Journals}

\maketitle
	
\begin{abstract}
Effective fusion of different types of features is the key to salient object detection.
The majority of existing network structure design is based on the subjective experience of scholars and the process of feature fusion does not consider the relationship between the fused features and highest-level features.
In this paper, we focus on the feature relationship and propose a novel global attention unit, which we term the "perception-and-regulation" (PR) block, that adaptively regulates the feature fusion process by explicitly modeling interdependencies between features.
The perception part uses the structure of fully-connected layers in classification networks to learn the size and shape of objects.
The regulation part selectively strengthens and weakens the features to be fused.
An imitating eye observation module (IEO) is further employed for improving the global perception ability of the network.
The imitation of foveal vision and peripheral vision enables IEO to scrutinize highly detailed objects and to organize the broad spatial scene to better segment objects.
Sufficient experiments conducted on SOD datasets demonstrate that the proposed method performs favorably against 22 state-of-the-art methods.
\end{abstract}


% \fi
\IEEEpeerreviewmaketitle

\section{Introduction}
\IEEEPARstart{S}{alient} object detection (SOD)~\cite{wang2019sodsurvey}\cite{2021-TMM-cmSalGAN}\cite{2021-TPAMI-rethinking-Co-SOD}\cite{2021-TNNLS-RethinkingRGBD}\cite{2021-TPAMI-NoisyLabel}\cite{2021-PR-EFNet} aims to find salient areas in an image~\cite{2020-CVPR-Scribble}\cite{2021-AAAI-RD3D}\cite{2021-TPAMI-Uncertainty} or video~\cite{2018-TIP-VSOD-FCN}\cite{2018-TPAMI-SA-VOS} by using intelligent algorithms that mimic human visual characteristics.
% SOD \cite{1998-TPAMI-SVA}
% image~\cite{2019-NC-Deepside}
% video~\\cite{2018-PAMI-VSOD-SAV}
%  This is usually the initial step of many image comprehension and processing tasks
It has been used in many image comprehension and video processing tasks, such as photo cropping~\cite{2019-TPAMI-PhotoCropping}, image editing~\cite{2010-ACMTOG-RepFinder}, 4D saliency detection~\cite{2019-ICCV-LightField} photo composition~\cite{2009-ACMTOG-Sketch2Photo}, and target tracking~\cite{2009-CVPR-tracking-SBDT}.
% non-photo-realist rendering~\cite{2013-GM-rendering},
% image retrieval~\cite{2012-TIP-3Dretrieval},

With the development of deep neural networks, various network structures and novel convolution modules are designed to improve the segmentation effect.
% There are still two key issues that deserve attention:
The majority of salient object detection networks are based on the U-shaped network~\cite{2015-CVPR-FCN} to integrate the features of different depths and scales.
The network structure represented by U-net~\cite{2015-ICM-Unet} and feature pyramid network (FPN)~\cite{2027-CVPR-FPNdetection}  has an obvious problem of semantic information dilution.
Therefore, transferring rich semantic information to shallow layers without losing location information and destroying details is the focus of current algorithms~\cite{2019-TPAMI-DSS}\cite{2020-AAAI-F3}\cite{2019-ICCV-EGNet}\cite{2019-CVPR-PFANet}\cite{2019-CVPR-PoolNet}\cite{2020-AAAI-GCPANet}.
Among them, the global guidance structure (GGS) represented by~\cite{2019-CVPR-PoolNet}\cite{2020-AAAI-GCPANet}\cite{2021-AAAI-SCWS} is widely used.
The existing methods have great contributions to network structure and module optimization.
These designs are based on rich experimental attempts and the subjective experience of scholars.
Although great progress has been made at present, two key issues still are worthy of further study: how to regulate different types of features to complete better segmentation from the perspective of the overall need for weight regulation and how to better balance the ability to organize a broad spatial scene with the ability to scrutinize highly detailed objects. % 这里原本加了斜体 \textit{}

\begin{figure}[t]% The dilation operation bring the influence of weight of edge pixels to the surrounding pixels.
\centering  % fully-connected
\includegraphics[width=0.9\columnwidth]{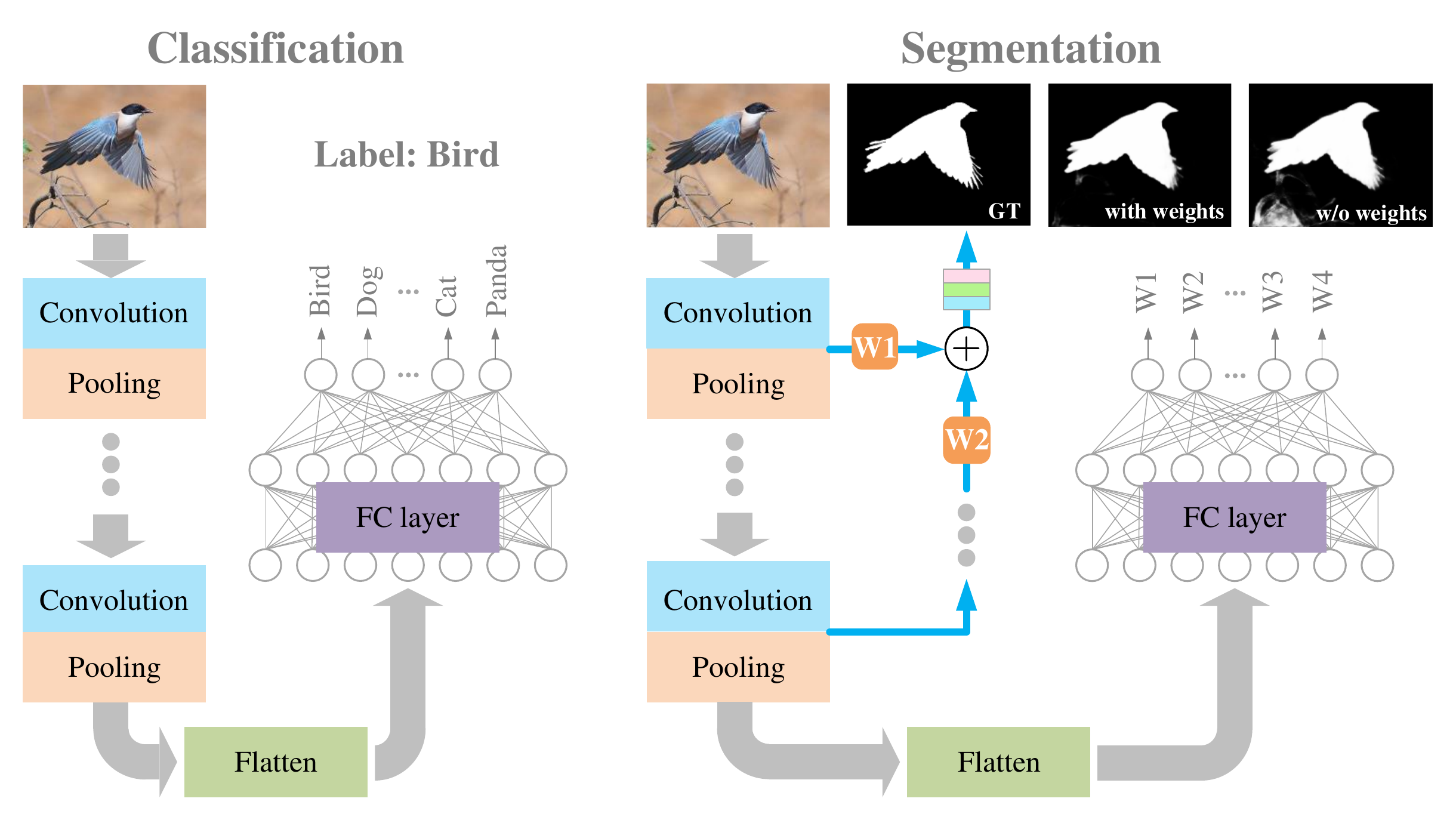} % Reduce the figure size so that it is slightly narrower than the column. Don't use precise values for figure width.This setup will avoid overfull boxes.
\caption{Inspired by the application of fully-connected layers in the classification networks, we replace the prediction of categories with the prediction of weights of features to be fused. The network regulated by weights has a better performance.}% Here $\lambda$ is 0.
\label{F1}
\end{figure}

Most of the existing methods fuse multi-level features directly without considering the contribution ratio of the fused features to the final output.
Therefore, for the first issue, we propose a perception-and-regulation (PR) block to optimize the FPN structure from the perspective of global perception and local feature fusion fine-tuning.
% The network structure has been greatly improved, we think that there is still room for improvement from the perspective of global perception and local feature fusion fine-tuning.
% Because we need local feature fusion with global perception to further optimize the network structure design, rather than rigidly fusing features directly.
Global perception helps to provide accurate semantic information and make better weight regulation,
and local feature fusion fine-tuning helps to enhance useful information and suppress invalid information.
The perception part of the PR block realizes global perception, which adopts the deepest feature with the largest receptive field as the input.
The regulation part of the PR block realizes local feature fusion fine-tuning, which adopts a weighted method to optimize the feature fusion process.

% For the second issue, many algorithms~\cite{2020-CVPR-COD}\cite{2019-CVPR-PFANet}\cite{2018-TPAMI-DeepLab} use atrous convolution to expand the perception range.
For the second issue, \cite{2020-CVPR-COD}\cite{2019-CVPR-PFANet}\cite{2018-TPAMI-DeepLab} use atrous convolution to expand the perception range.
The too-large atrous ratio will make the information at independent sampling points discontinuous, which is not conducive to the continuity of spatial information in detail.
Therefore, the convolution combination of multiple atrous rates is widely adopted~\cite{2019-CVPR-PFANet}\cite{2018-TPAMI-DeepLab}\cite{2020-arxiv-detectRS} to make the network have the ability to organize a wide space scene and scrutinize highly detailed objects.
To solve this problem, we propose our solution, the imitating eye observation module (IEO), to better balance the two abilities.

\begin{figure*}[htb]% The dilation operation bring the influence of weight of edge pixels to the surrounding pixels.
\centering
\includegraphics[width=2\columnwidth]{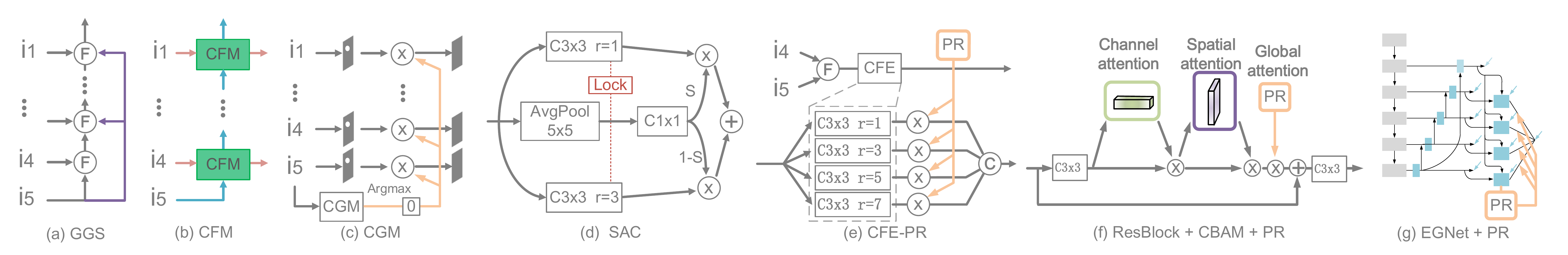} % Reduce the figure size so that it is slightly narrower than the column. Don't use precise values for figure width.This setup will avoid overfull boxes.
\caption{Related module diagrams and the application of PR block in CFE, CBAM, and EGNet. (a) Simplified global guidance structure(GGS)~\cite{2020-AAAI-GCPANet}~\cite{2019-CVPR-PoolNet}. (b) Cross feature module (CFM) of ~\cite{2020-AAAI-F3} (a) (b) is used for semantic information enhancement. (c) Classification-guided module (CGM) of Unet3+~\cite{2020-ICASSP-UNet3+} is used to judge whether it is noise or organ. (d) Switchable trous revolution (SAC)~\cite{2020-arxiv-detectRS} is helpful to detect objects of different scales. (e) Context-aware feature extraction module~\cite{2019-CVPR-PFANet} with PR block (f) Convolutional block attention module (CBAM)~\cite{2018-ECCV-CBAM} with PR block (g) Simplified structure of EGNet~\cite{2019-ICCV-EGNet} with PR block.}% Here $\lambda$ is 0.
\label{related}
\end{figure*}

A partition search strategy is adopted in IEO to effectively alleviate the negative impact of a too-large atrous ratio.
The adoption of the PR block enables the IEO module to further balance different types of features in IEO, which helps to further improve network performance.
%But for different types of objects, how to balance different types of features to achieve better segmentation has become a new problem.

Inspired by squeeze-and-excitation (SE) block~\cite{2020-TPAMI-SENet} and the classification network, we design a global attention unit called perception-and-regulation (PR) block to regulate the network automatically and achieve the best feature fusion effect.
Our goal is to improve the feature fusion effect by explicitly modeling the interdependencies between features to be fused.
To achieve this, we design a mechanism that helps the network to recalibrate the feature fusion process, through which it can learn to use global information to adaptively adjust the weights of the features to be fused.
In addition, we design an imitating human eye observation module (IEO) to organize a broad spatial scene and have the ability to scrutinize highly detailed objects.
% Our model only uses 130M to achieve excellent performance.
Our contributions are summarized as follows:
\begin{itemize}
% ***attention***
\item We propose a perception-and-regulation (PR) block to help the network understand the global information and assign the feature weights uniformly to realize the spontaneous and adaptive global feature regulation.
\item An imitating human eye observation module (IEO) is proposed to help the network have the ability to organize a wide space scene and scrutinize highly detailed objects. PR block balances and optimizes these two abilities.
% \item We utilize a boundary pixel perception loss to allocate training weights appropriately for boundary pixels. Corner perception ability of BPP loss helps the model obtain accurate local details.
\item Sufficient experiments conducted on 5 SOD datasets demonstrate that the proposed method outperforms 22 state-of-the-art SOD methods in terms of eight metrics. In addition, ablation experiments on several modules and networks prove the universality of our PR block.

\end{itemize}

\section{Related work}

\subsection{Salient object detection}
% => DEL1
%\textcolor{blue}{Image saliency is a classic problem which is widely concerned for decades, and it is also the basic step of many image processing tasks.
%The related research fields include video SOD~\cite{2018-TIP-VSOD-FCN}\cite{2018-PAMI-VSOD-SAV}, light field SOD~\cite{2017-TPAMI-LightField}, Co-SOD~\cite{2021-TPAMI-rethinking-Co-SOD}, RGB-D SOD~\cite{2021-PAMI-UCNet}\cite{2021-AAAI-RD3D}, weakly supervised SOD~\cite{2020-CVPR-Scribble}, unsupervised SOD~\cite{2018-CVPR-DUSD}, etc.
%The prosperity of these research fields allows SOD to be better applied in more scenarios.}
Early salient object detection methods are based on hand-crafted features~\cite{2009-CVPR-Frequency}\cite{2015-TPAMI-GC}\cite{2013-CVPR-Submodular}\cite{2011-ICCV-Center}\cite{2013-CVPR-Discriminativ}\cite{2013-CVPR-Hierarchical}\cite{2013-CVPR-Graph}\cite{2016-TIP-CDST} and intrinsic cues without a deep hierarchical structure.
With the development of deep learning, the deep features with rich contextual information make a great breakthrough in the field of salient object detection.
As a landmark algorithm, fully convolutional networks (FCN)~\cite{2015-CVPR-FCN} creatively removes the fully-connected layer to predict the semantic label for each pixel.
Then the U-shape based structures represented by U-net~\cite{2015-ICM-Unet} and feature pyramid network (FPN)~\cite{2027-CVPR-FPNdetection} has gradually become the mainstream structure by integrating all levels of features layer by layer.
Based on this structure, scholars have explored more abundant multi-layer feature blending methods to help feature expression effectively. Among them, the global guidance structure (GGS) represented by~\cite{2019-TPAMI-DSS}\cite{2020-AAAI-GCPANet}\cite{2019-CVPR-PoolNet} has gradually become a common method to strengthen semantic information.

Our $\emph{Perception-and-Regulation}$ (PR) network is based on GSS to regulate the network spontaneously and adaptively.
It is worth noting that, in contrast to FCN abandoning the fully-connected layer (FC), our network mainly relies on the FC layer in the classification network to perceive the size and shape of objects. Then the weights of the features to be fused are evaluated according to the semantic information obtained from the FC layer, as shown in Fig.\ref{F1}.

\subsection{Semantic information reinforcement} % Global guidance structure.
Scholars have proposed various methods to transmit global information of high-level features to the shallow layers to help the network get detail information and accurately locate the objects.
For example, Hou \textit{et al.}~\cite{2019-TPAMI-DSS} proposed short connections to help shallower side-output layers get semantic information more directly. % shallower side-output layers better locate the salient region.
Zhao \textit{et al.}~\cite{2019-ICCV-EGNet} used the highest-level features to help the edge features (shallow layer) of explicit modeling to filter out useless edge details.
Zhao \textit{et al.}~\cite{2019-CVPR-PFANet} designed a spatial attention module where context-aware high-level features are added to help the location information transfer to the shallow layer. %  after channel-wise attention
Liu \textit{et al.}~\cite{2019-CVPR-PoolNet} introduced a global guidance module (GGM) to explicitly make shallower layers be aware of the locations of the objects.
Global context flow module in ~\cite{2020-AAAI-GCPANet} solved the issue of dilution in the process of high-level feature transmission, which is similar to GGM.

\begin{figure*}[htb]% The dilation operation bring the influence of weight of edge pixels to the surrounding pixels.
\centering
\includegraphics[width=1.9\columnwidth]{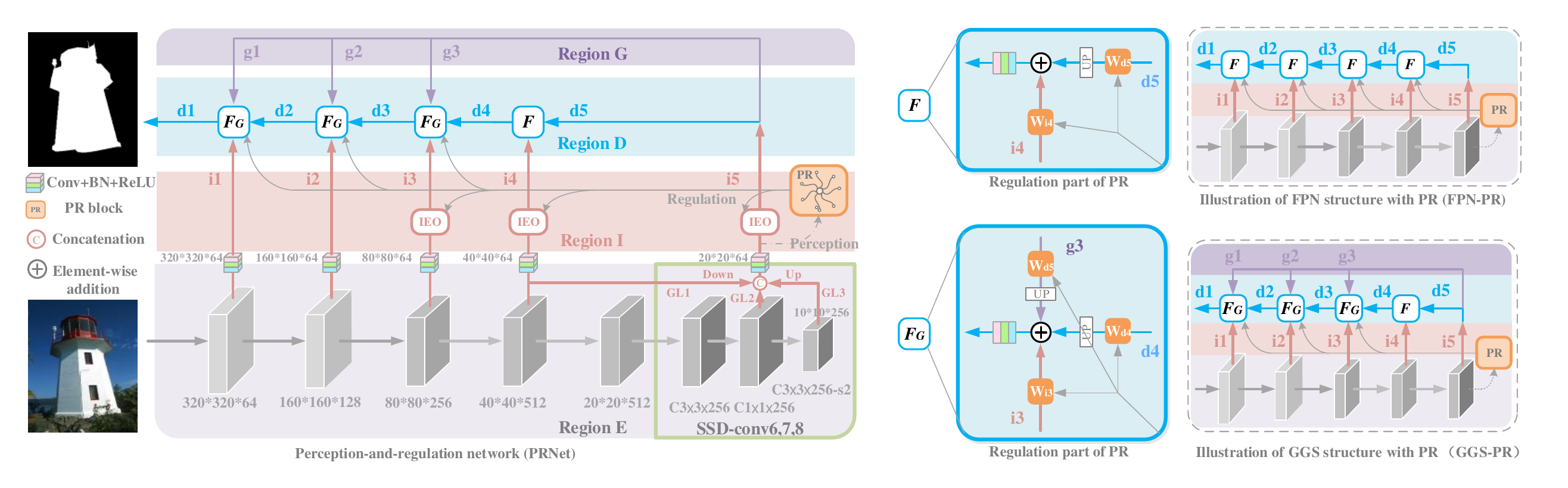} % Reduce the figure size so that it is slightly narrower than the column. Don't use precise values for figure width.This setup will avoid overfull boxes.
\caption{Perception-and-regulation network (PRNet).
On the right side of the picture are the applications of PR block in FPN structure and GGS structure.
The middle part of the picture shows the regulation part of PR (F-PR and F$_{G}$-PR).
Besides, F in the GGS structure is not regulated by PR.
% On the left side of the picture, IEO modules are added on the basis of GSS-PR.
On the left side of the picture, we add SSD structure and IEO modules to GGS-PR to get the final network PRNet.
i1, i2, i3, i4, i5 in region I are the interlayer features.
d1, d2, d3, d4, d5 in region D are the decoder features.
g1, g2, g3 in region G are global guidance features.
The gray dotted arrow represents the input feature of the perception part of PR block.
The gray solid arrows in region I represent the regulation of the output weights of the PR block on each feature fusion process.
}% Here $\lambda$ is 0.
\label{pipline}
\end{figure*}

The global guidance idea of~\cite{2020-AAAI-GCPANet}~\cite{2019-CVPR-PoolNet} can be simplified to GGS structure in Fig.~\ref{related} (a) and the structure can effectively solve the problem that semantic information is diluted in the process of feature transfer of FPN structure or U-net structure~\cite{2015-ICM-Unet}.
Even if these structural designs are based on the subjective experience and repeated attempts~\cite{2019-TPAMI-DSS} of excellent scholars, the rigid feature fusion process does not consider the relationship between different features to be fused and the highest level features.
The GGS-PR structure of Fig.~\ref{pipline} uses PR block to further slightly regulate and optimize the network.
Wei \textit{et al.}~\cite{2020-AAAI-F3} designed a cross feature module (CFM) to select features with rich semantic information to transfer to
the shallower layer and let the features with details enter the next cycle as shown in Fig.~\ref{related} (b).
CFM can be understood as strengthening the transmission of semantic information in FPN structure.
Our PR block does the same work, but the way of strengthening semantic features is more flexible and adaptive as shown in the FPN-PR of Fig.~\ref{pipline}.

\subsection{Attention mechanisms}
In the classification task, Hu \textit{et al.}~\cite{2020-TPAMI-SENet} improves the quality of feature representation greatly by establishing the interdependencies between the channels of its convolutional features.
Correspondingly, Woo \textit{et al.}~\cite{2018-arXiv-BAM}\cite{2018-ECCV-CBAM} use the spatial attention map generated by utilizing the inter-spatial relationship of features with the channel attention map to help the network learn 'where' is an informative part and 'what' is meaningful on the spatial and channel axis respectively.
Some SOD methods adopt attention mechanism~\cite{2020-TIP-RAS}\cite{2018-CVPR-PAGRN}\cite{2019-CVPR-CPD}\cite{2019-CVPR-PFANet}.
Wang \textit{et al.}~\cite{2019-CVPR-PAGE} design the pyramid attention module to make the network pay more attention to important regions and multi-scale information.
As a special attention mechanism, the gated mechanism is widely used by long short term memory (LSTM) and gated recurrent unit (GRU), which play an important role in SOD algorithms~\cite{2019-CVPR-TopDown}\cite{2019-PAMI-ASNet}\cite{2019-CVPR-CapSal}.
Some segmentation algorithms~\cite{2017-CVPR-GatedDense}\cite{2018-CVPR-BMPM}\cite{2020-ECCV-GateNet} use the gated mechanism to adjust the network.
CGM of Unet3+~\cite{2020-ICASSP-UNet3+} can be regarded as the extreme case of the gated mechanism, because the weights of controlled features are set to 0 or 1 via argmax (Fig.\ref{related}(c)), which helps to determine whether the target is an organ or noise.

Inspired by SENet~\cite{2020-TPAMI-SENet} and the classification network, we design a PR block for global regulation which can be regarded as a macro global attention mechanism.
\cite{2020-TPAMI-SENet} adaptively recalibrates channel-wise features at the micro level, while PR block recalibrates different types of features of the whole network at the macro level.
Different from the existing methods of the gated network, all the features to be fused are regulated in our network and the perception part is located in the position with rich semantic information, which helps to analyze the size and shape of the object accurately and uniformly.
In addition, inspired by SAC of~\cite{2020-arxiv-detectRS} (Fig.\ref{related}(d)), we use softmax, which is commonly used in classification networks, to add constraints to the regulation part of the PR block.

Some algorithms~\cite{2019-CVPR-PFANet}\cite{2020-CVPR-COD}\cite{2020-arxiv-detectRS}\cite{2020-ECCV-GateNet}\cite{2018-TPAMI-DeepLab} use the atrous convolution to expand the receptive range to better observe the object.
The disadvantage of atrous convolution with a large atrous ratio is that the information given by spatial continuity may be lost (such as edges) and it is not conducive to the segmentation of small objects.
We design a special spatial attention mechanism to compensate for this shortcoming by imitating foveal vision and peripheral vision. % (Fig.\ref{IEO})

\section{Perception-and-Regulation Block}
The $\emph{Perception-and-Regulation}$ (PR) block is a computational unit with semantic perception and global regulation capability.
The PR block serves the feature level weight regulation of our final network PRNet, as shown in the network structure on the left side of Fig.\ref{pipline}.
Besides, it can also be used in many network structures or modules.
% =>DEL2
%\textcolor{blue}{PRNet is composed of GGS structure, PR block and IEO modules.
%So we first introduce three perception strategies of PR in Sec.3.1.
%Then we analyze the design of the regulation part of PR block for FPN structure, GGS structure, CBAM module (Fig.\ref{related} (f))~\cite{2018-ECCV-CBAM}, and CFE module (Fig.\ref{related} (e))~\cite{2019-CVPR-PFANet} in Sec.3.2.
% The purpose of the above design is to verify the universality of the PR block and pave the way for the IEO module we proposed in Sec.4.
%Because PRNet is based on GGS-PR structure, please refer to Sec.3.2.1 for specific feature name definition.
% We begin with the PR block.}

Inspired by SENet~\cite{2020-TPAMI-SENet} and the classification network, we design a PR block to make the fusion process of different types of features adaptively adjust according to the overall need of weight regulation.
SE block focuses on features and adaptively recalibrates features at the channel level, while our PR block focuses on the entire network and recalibrates the entire network at the feature level.
% Fig.\ref{PRblock} shows this relationship.
Therefore, PR block can be considered as a macro version of SE block.
% => DEL3
% \textcolor{blue}{The specific differences between SE block and PR block are as follows:
% 1. The calibration target of SE block is feature.
% However, the calibration target of PR block is the whole network.
% 2. SE block is weighted for the interior channels of feature and recalibrated for the relationship between channels.
% But PR block is weighted for the features of network and recalibrated for the relationship between features.
% 3. The input of the SE block is an uncalibrated feature and its weighed position is the channels of the uncalibrated feature.
% But the input of PR block is a feature with rich semantic information and its weighted position are the uncalibrated features.}

In addition, FCN~\cite{2015-CVPR-FCN} adapts classification networks (AlexNet~\cite{2012-NIPS-AlexNet}, VGG net~\cite{2015-ICLR-VGG} and GoogLeNet ~\cite{2015-CVPR-GoogLeNet}) into fully convolution networks by replacing the fully-connected (FC) layer with convolution layers to achieve semantic segmentation.
On the contrary, we make full use of the perception and understanding ability of the FC layer to make adaptive regulation for the whole network.
In Sec.3.1, we discuss three perception strategies in detail for the perception part of the PR block.
PR block solves the first issue of how to regulate different types of features to complete better segmentation
\begin{figure*}[htb]% The dilation operation bring the influence of weight of edge pixels to the surrounding pixels.  senet6
\centering
\includegraphics[width=1.8\columnwidth]{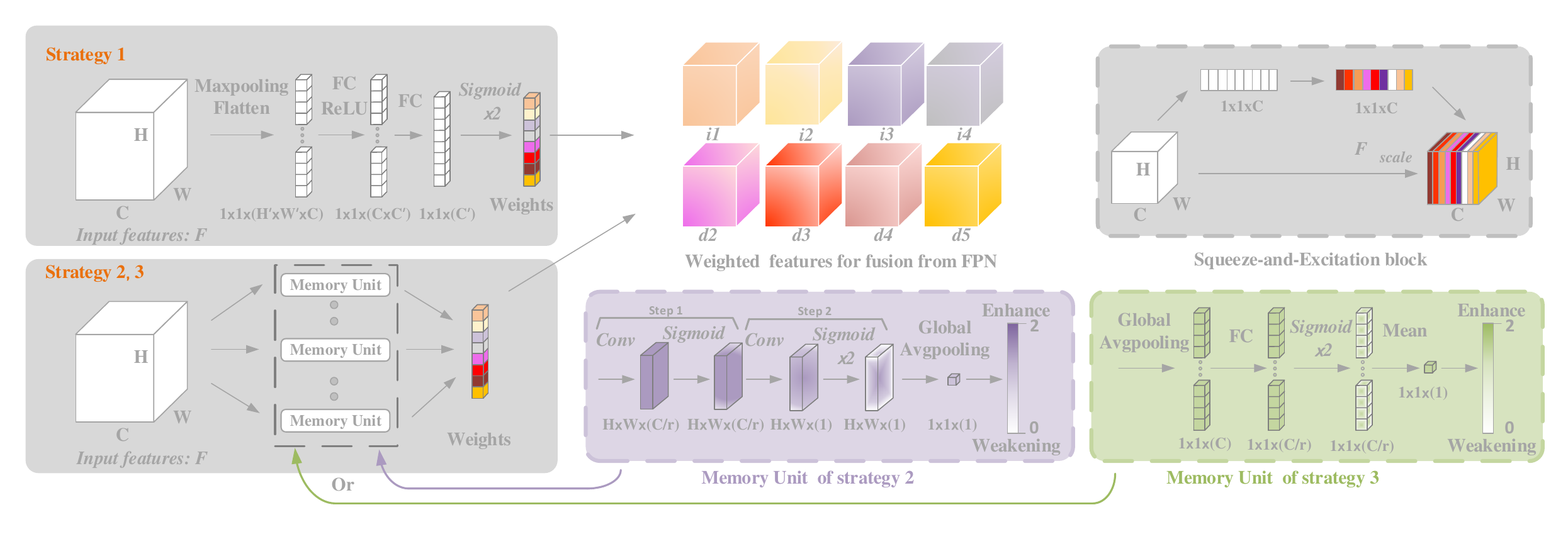} % Reduce the figure size so that it is slightly narrower than the column. Don't use precise values for figure width.This setup will avoid overfull boxes.
\caption{Three perception strategies for the perception-and-regulation block.
Strategy 1 uses the fully-connected layer of classification networks to evaluate the weights of the features to be fused.
Strategy 2,3 generate feature weights independently for each feature to be fused according to the highest level feature.
The dimension change of the weight of strategy 2 is (HxWxC)-(HxWx1)-(1x1x1).
The process of strategy 3 is (HxWxC)-(1x1xC)-(1x1x1).
The features i1, i2, i3, i4, d2, d3, d4, d5 to be regulated come from the interlayer region (Region I) and decoder region (Region D) of FPN, which correspond to the upper right corner of Fig.\ref{pipline}.
For the convenience of comparison, we show the SE block in the upper right corner~\cite{2020-TPAMI-SENet}.
}% Here $\lambda$ is 0.
\label{PRblock}
\end{figure*}
from the perspective of the overall need for weight regulation.
PR block has a better regulation effect on the fusion process of features with large differences.
The differences of features here refer to different levels of features (FPN GGS), the features produced by atrous convolution or normal convolution (CFE), the original features or attention features in the residual structure of the attention module (CBAM).

% In Sec.3.1, three perception strategies of perception part are introduced.
% In Sec.3.2, we introduce the design of the regulation part for FPN structure, GGS structure, CBAM module (Fig.\ref{related} (f))~\cite{2018-ECCV-CBAM} and CFE module (Fig.\ref{related} (e))~\cite{2019-CVPR-PFANet}.
% The purpose of the above design is to verify the universality of the PR block and pave the way for the IEO module we designed in Sec.4.

\subsection{Perception: Semantic Information Embedding}

In the perception part of the PR block, three semantic information embedding methods are designed (Fig.\ref{PRblock}) according to the idea of SENet for global information embedding.
In order to introduce our idea clearly, we annotate the features in FPN-PR in the upper right corner of Fig.\ref{pipline}.
% DEL
% \textcolor{blue}{The encoder is in the gray area, the interlayer features (i1, i2, i3, i4, i5) are in the red area and the decoder features (d1, d2, d3, d4, d5) are in the blue area.}
%The weighted features in Fig.\ref{PRblock} refer to the features in Fig.\ref{pipline} FPN-PR.

$\emph{Strategy 1}$ in Fig.\ref{PRblock} uses the FC layer to perceive the size and shape of the objects.
The output of the FC layer of the classification network is the category of the predicted object, while our output results are the weights of the features to be fused.
The weights are adjusted adaptively according to the characteristics of the objects.
Features regulated by weights are represented by the color of the corresponding weight.
Unlike the feature weighting of the SE block for a single channel (Fig.\ref{PRblock}), the PR block regulates the whole features.
The perception part of $\emph{strategy 1}$ is set at the location of high-level features (i5) with rich semantic information.
Max-pooling is used to reduce the size of the feature $\mathbf{F}$$\in$$\mathbb{R}^{H \times W \times C}$.
Then the feature $\mathbf{F_{m}}$$\in$$\mathbb{R}^{H^{'} \times W^{'} \times C}$ after pooling is transformed into a one-dimensional vector $\mathbf{F_{f}}$$\in$$\mathbb{R}^{1\times 1 \times ( H^{'} \times W^{'} \times C)}$ by flatten operation.
We use a multi-layer perceptron (MLP)~\cite{2018-arXiv-BAM} with one hidden layer to enhance the perception ability of the network.
To save a parameter overhead, the hidden activation size is set to $\mathbb{R}^{1 \times 1 \times (C \times C^{'})}$.
C$^{'}$ is the number of features that need to be regulated and it is eight for the FPN structure in Fig.\ref{PRblock}.
The output layer size is $\mathbb{R}^{1 \times 1 \times (C^{'})}$.
The activation function of the output layer replaces ReLU with sigmoid and the final results are multiplied by 2 to show whether the features are enhanced or suppressed.
In short, the perception part (\textbf{P}$_{FC}$) is computed as:

\begin{small}
\begin{equation}\label{map}
\mathbf{P}_{FC} = 2*(MLP_{S}(Flatten((MaxPool(\mathbf{F}))))).
\end{equation}
\end{small}

The activation function of $MLP_{S}$ output layer is element-wise sigmoid.

Due to the dense connection of FC layers, the final weights of $\emph{strategy 1}$ have a strong correlation.
In order to explain the perception part of the PR block more clearly, we provide two other spatial dimension (S) and channel dimension (C) perception design.
Different from $\emph{strategy 1}$, we design multiple independent memory units (MU) for $\emph{strategy 2,3}$ (Fig.\ref{PRblock}) to better illustrate the mapping relationship.
The function of MU is to establish the mapping relationship between the shape and size of input features $\mathbf{F}$ and the weight of regulated features.
The number of MU is determined by the number of features that need to be regulated.
In the MU of $\emph{strategy 2}$ , we use two convolution operations to gradually reduce the channel dimension of input features to 1.
Here we refer to the design of spatial attention map~\cite{2018-arXiv-BAM}\cite{2018-ECCV-CBAM}.
r is the reduction ratio.
The purpose of adding convolution here is to get the output weight according to the input features, and it can be understood that the convolutions change the average gray value of the input features.
The weight is equal to the average of the output feature (channel is 1).
We change the activation function to sigmoid to enhance the nonlinearity of the output and multiply the final result by 2.
In short, the perception part (\textbf{P}$_{S}$) is computed as:

\begin{small}
\begin{equation}\label{map}
\mathbf{P}_{S} = G(2* C_{S}(C_{S}(\mathbf{F}))),
\end{equation}
\end{small}
where $C_{S}(\cdot)$ is the convolution operation using the element-wise sigmoid function and $G(\cdot)$ is th global average pooling.

For the MU of $\emph{strategy 3}$, we use global average pooling to reduce the input feature to one dimension.
Then we use the fully-connected layer to reduce the channel to $C/r$.
After sigmoid and average operation, the final weight is obtained.
The perception part (\textbf{P}$_{C}$) is computed as:

\begin{small}
\begin{align}\label{map}
\mathbf{P}_{C} = Avg(2* FC_{S}(G(\mathbf{F}))),
\end{align}
\end{small}
where $FC_{S}(\cdot)$ is the fully-connected layer and the activation function of the output layer is sigmoid.  % element-wise sigmoid
$Avg(\cdot)$ is the average operation of the one-dimensional vector.

\subsection{Regulation: Adaptive Recalibration}
The accurate location information of high-level features in FPN structure is diluted in the process of multiple fusion~\cite{2020-AAAI-GCPANet}.
This is because element-wise addition or concatenation operations do not treat the weights of the features to be fused differently.
Most of the current algorithms focus on changing the network structure~\cite{2020-AAAI-GCPANet}\cite{2019-ICCV-EGNet}\cite{2019-CVPR-PoolNet}\cite{2019-TPAMI-DSS} or module~\cite{2020-AAAI-F3}\cite{2019-CVPR-PAGE} to enhance semantic information, while the PR network focuses on the regulation of network and can greatly improve the performance of simple network structure.
The PR network uses a PR block for global regulation.
It builds a bridge between the features to be fused and the semantic information made by the fully-connected layer.
The semantic information is expressed in the form of weight.

\subsubsection{Basic Structural Analysis (FPN-PR and GGS-PR)}

Both \cite{2019-CVPR-PoolNet} and \cite{2020-AAAI-GCPANet} use the global guidance method to enhance the semantic information of the shallow features of the network.
We add a global guidance structure to the FPN to imitate this process, as shown in Fig.\ref{related} (a).
The global guidance structure can be considered as a simplified version of ~\cite{2019-TPAMI-DSS}\cite{2019-CVPR-PoolNet}\cite{2020-AAAI-GCPANet}.
The location features are directly added to the shallow features to enhance the ability of salient object location.
% There are many structural designs of this type, which are mainly based on subjective design and multiple experimental modifications.
There is still room for improvement.
We add PR block to both FPN and GGS structures (Fig.\ref{pipline}) for further slight regulation.
\cite{2020-AAAI-F3} proposes a CFM module to help the output features with high-level features (Fig.\ref{related} (b) blue arrow) as the main components to transfer to the shallow layer.
This scheme is not flexible enough because the structure of CFM is fixed.
While the perception part of the PR block helps us to adaptively recalibrate the weight of each feature in the whole network according to the characteristics of the object to achieve better segmentation.

Fig.\ref{pipline} shows the realization of perception and regulation of PR block.
In order to simplify the network and reduce the amount of computation, we use convolution (convolution, batch normalization, ReLU) to unify the output features of the encoder to 64 channels.
FPN-PR and GGS-PR on the right side of Fig.\ref{pipline} do not show this detail.
Taking FPN-PR with perception $\emph{strategy 2}$ as an example, eight memory units are used to perceive the input feature and evaluate the weights of interlayer features (i1, i2, i3, i4, i5) and decoder features (d1, d2, d3, d4, d5).
The gray dotted arrow represents the input feature $\mathbf{F}$ of the perception part.
The gray solid arrows represent the regulation of the eight output weights of the PR block on each feature to be fused.
The only difference from the traditional FPN structure is that PR block weights these features.
The features (g1, g2, g3) in the purple region G are global guidance features.
It is worth noting that we only weighted the three feature fusion process of GGS to explore the influence of PR block on the global guidance.
In short, the FPN-PR and GGS-PR are computed as:

\begin{scriptsize}
\begin{align}\label{E-GSS-FPN}
d_{j-1}^{FPN} &= C(W_{i_{j-1}}* i_{j-1}+U(W_{d_{j}}* d_{j})),\\
d_{j-1}^{GGS} &= C(W_{i_{j-1}}* i_{j-1}+U(W_{d_{j}}* d_{j})+U(W_{g_{j-1}}* d_{j-1})),
\end{align}
\end{scriptsize}
where $C(\cdot)$ refers to the convolution operation. $W$ is the weight. $U(\cdot)$ is upsampling.

In order to make the perception part of PR block have a better perception effect on different scale objects, we adopt the partial encoder design of SSD algorithm~\cite{2016-ECCV-SSD}, as shown in the left part of Fig.\ref{pipline}.
The advantage of SSD is that it can detect objects on multiple scales.
We use its 6,7,8 layer structure and combine 4,7,8 layer features with concatenation to realize multi-layer perception.
% We will describe the convolution parameters and the features size in detail in the attachment.

%The convolutions added after VGG are 3x3 conv$\in \mathbb{R}^{256}$, 1x1 conv$\in\mathbb{R}^{256}$, 3x3 conv$\in\mathbb{R}^{256}$.
%Because the last convolution step size is 2, so the features of layer 4$\in\mathbb{R}^{40\times 40\times 516}$ and layer 8$\in\mathbb{R}^{10\times 10\times 256}$ are merged with layer 7$\in \mathbb{R}^{20\times 20\times 256}$ by down sampling and up sampling. When the encoder features enter the region I, we unify the channels of all levels of features to 64.

\subsubsection{Exemplars (CFE-PR, CBAM-PR and EGNet-PR)}

In order to verify the universality of PR block, we apply it to context-aware feature extraction module (CFE) in Fig.\ref{related} (e)~\cite{2019-CVPR-PFANet} and CBAM module in Fig.\ref{related} (f)~\cite{2018-ECCV-CBAM}.
The features to be fused in FPN and GGS structures are features of different scales and depths, while the features to be fused in CFE-PR modules have different receptive fields.

Different from~\cite{2019-CVPR-PFANet}, there are all 3x3 convolution in CFE of this paper and their dilation rates are 1,3,5,7 respectively.
CFE is used in the feature i3, i4 and i5 position (follow~\cite{2019-CVPR-PFANet}) of basic FPN structure to enhance the output feature and PR block is used for internal regulation as shown in Fig.\ref{related} (e).

As complementary attention modules, channel-wise attention and spatial attention are used to calibrate features at the spatial and channel levels and they can learn 'what' and 'where' to attend in the channel and spatial axes respectively~\cite{2020-TPAMI-SENet}\cite{2018-arXiv-BAM}\cite{2018-ECCV-CBAM}.
PR block can be considered as the third type of attention, which focuses on the influence of the whole feature on the network (FPN, GGS) or module(CFE, CBAM).
The perception part of the PR block analyzes the global context information of high-level features, and strengthens or weakens the features to be fused from the perspective of the global needs of the network. So we call it global attention in Fig.\ref{related} (f).
We use PR block with channel attention and spatial attention to further optimize the attention mechanism.
The regulation part of the PR block is added to the attention branch of CBAM in Fig.~\ref{related} (f) and five CBAM-PR modules are added to i1, i2, i3, i4, i5 positions of the basic FPN structure.

We also added PR block to the final output position of EGNet~\cite{2019-ICCV-EGNet} for perception and regulation, and the final result was further improved. Fig.\ref{related} (g) is a simplified structure diagram of EGNet, and we added PR block in its final output location (FPN structure).
% We will introduce the specific location of the perception and regulation parts in the attachment.
Line 11-16 of Tab.\ref{FPN} proves the effect of PR block in the above modules and networks.

\begin{figure*}[htb]% The dilation operation bring the influence of weight of edge pixels to the surrounding pixels.
\centering
\includegraphics[width=1.9\columnwidth]{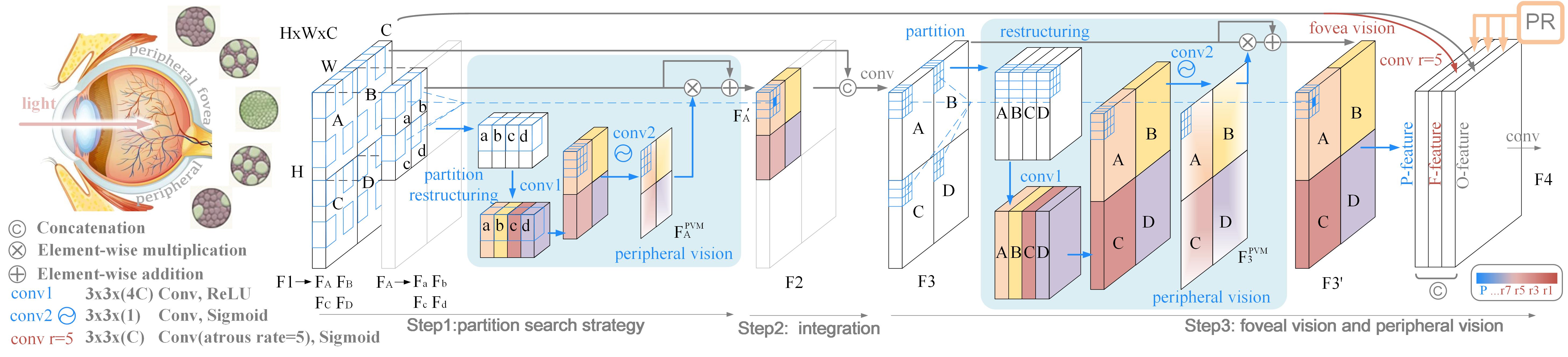} % Reduce the figure size so that it is slightly narrower than the column. Don't use precise values for figure width.This setup will avoid overfull boxes.
\caption{Imitating human eye observation module.
The blue area is the peripheral vision module, which is applied twice in steps 1, 3 to make the peripheral vision feature quickly obtain the global receptive field. Under the regulation of PR block, peripheral vision feature, fovea vision feature, and original feature are fused.
}% Here $\lambda$ is 0.
\label{IEO}
\end{figure*}

\section{Imitating Eye Observation Module}
The purpose of the imitating human eye observation module (IEO) is to quickly and accurately find and locate salient objects.
IEO uses a partition search strategy (Fig.\ref{IEO} step1) to enable the network to focus on the analysis of local areas. %(F1$\rightarrow$F2 in Fig.\ref{IEO-detail})
Then it uses integration operation (Fig.\ref{IEO} step2) to associate the local feature analysis results with global information.
Inspired by human perception of foveal vision and peripheral vision~\cite{PV1}, we propose a peripheral vision module (PVM) (Fig.\ref{IEO} blue region) to cooperate with the foveal vision to achieve accurate understanding of the salient objects in step3.
The peripheral vision analysis strategy of PVM is applied in both the regional integration step1 and the global search (step3).
Step2 is a bridge to make the perception range of peripheral vision wider.

\begin{figure}[t]% The dilation operation bring the influence of weight of edge pixels to the surrounding pixels.
\centering
\includegraphics[width=0.9\columnwidth]{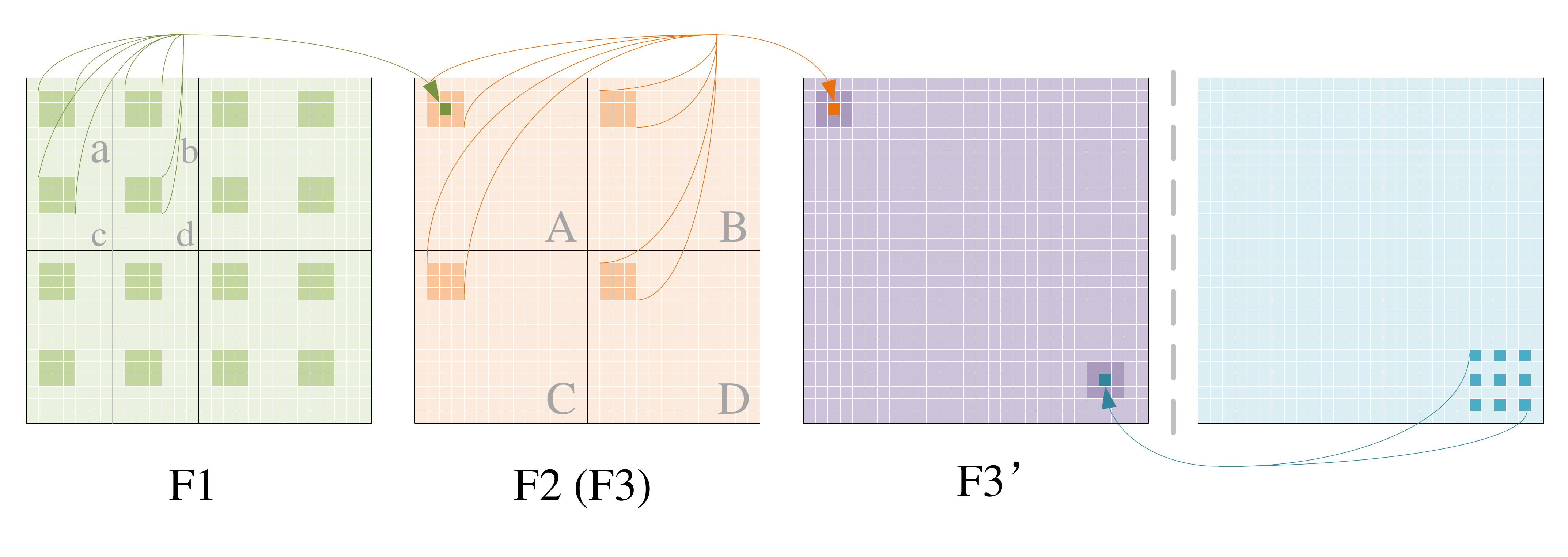} % Reduce the figure size so that it is slightly narrower than the column. Don't use precise values for figure width.This setup will avoid overfull boxes.
\caption{The receptive field of features in peripheral vision module (PVM).
The green, orange, and purple squares represent the features F1, F2(F3), and F3' in Fig.\ref{IEO}, respectively.
The receptive fields of the green pixel in F2 and the orange pixel in F3 'are indicated by the guide arrows respectively.
The receptive field of the orange pixel in F3' can cover the dark green areas in F1.
The atrous convolution (blue square on the far right) is used to compare with PVM.
}% Here $\lambda$ is 0.
\label{IEO-detail}
\end{figure}

In $\emph{step 1}$, we divide the input feature $\mathbf{F1}$ into four regions in the spatial dimension ($Par_{S}$) and analyze each region separately with PVM.
This mimics the human partition search strategy, which helps to find objects that are not in the center of view.
In addition, this operation also helps to form the global receptive field (we will explain it in detail in step 3).
Because the sub-regions $F_{A}, F_{B}, F_{C}, F_{D}$ of $\mathbf{F1}$ are partitioned again and then they all pass through the PVM module, we only show the implementation details of $F_{A}$ in Fig.\ref{IEO} step1 and Eq.7-10.
The features $F_{a}, F_{b}, F_{c}, F_{d}$ obtained by partitioning $F_{A}$ in spatial dimension ($Par_{s}$) are reorganized in channel dimension ($Cat_{c}$) in Eq.8.

\begin{footnotesize}
\begin{align}\label{step1}
&F_{A}, F_{B}, F_{C}, F_{D}  = Par_{s}(\mathbf{F1}),& \\
&F_{a}, F_{b}, F_{b}, F_{d}  = Par_{s}(F_{A}),&\\
&F_{A}^{PVM} = C^{2}(Cat_{s}(Par_{c}(C^{1}(Cat_{c}(F_{a}, F_{b}, F_{b}, F_{d}))))),&\\
&F_{A}^{'} = F_{A} + F_{A}*F_{A}^{PVM}, &
\end{align}
\end{footnotesize}
where $Par_{s}(\cdot)$, $Par_{c}(\cdot)$, $Cat_{s}(\cdot)$, $Cat_{c}(\cdot)$ refer to spatial dimension partition, channel dimension partition, spatial dimension concatenation, channel dimension concatenation. $C^{1}(\cdot) \in \mathbb{R}^{4C}$ is 3x3 convolution 1 in PVM and activated with ReLU. $C^{2}_{S}(\cdot) \in \mathbb{R}^{1}$ is 3x3 convolution 2 in PVM and activated with sigmoid.

In $\emph{step 2}$, we merge the results of partition search (Eq.10) in spatial dimension to get $\mathbf{F2}$ and then use concatenation operation to merge $\mathbf{F1}$, $\mathbf{F2}$ in channel dimension.
$\emph{Step 2}$ strengthens the association between the region search feature and the original feature.

\begin{footnotesize}
\begin{align}\label{step2}
&\mathbf{F2} = Cat_{s}(F_{A}^{'}, F_{B}^{'}, F_{C}^{'}, F_{D}^{'}),&\\
&\mathbf{F3} = C(Cat_{c}(\mathbf{F1}, \mathbf{F2})),&
\end{align}
\end{footnotesize}
where $C(\cdot)$ $\in \mathbb{R}^{C}$ .

In $\emph{step 3}$, we do the peripheral visual perception of $\mathbf{F3}$ again.
This process is the same as the previous Eq.7-9.
PVM can be considered as a special attention mechanism.
It expands the receptive field by comparing the features of corresponding positions in other partitions and then corrects the original features in spatial dimension (F1$\rightarrow$F2 in Fig.\ref{IEO-detail}).
Eq.9 shows the combination of attention branch and primitive branch~\cite{2018-ECCV-CBAM}\cite{2018-arXiv-BAM}.
PVM can also be considered as a special atrous convolution and has only \textit{four sampling points} and a very \textit{large atrous rate} (Fig.\ref{IEO-detail} F3$\rightarrow$F3$^{'}$).
The PVM here needs to be used for the feature of a large receptive field, otherwise, the spatial information discontinuity caused by a large atrous rate will appear.
% The problem of too-large void ratio can be solved by multiple partition and integration operations (F1$\rightarrow$F2$\rightarrow$F3).
The problem can be solved by partitioning and stacking PVMs (F1$\rightarrow$F2$\rightarrow$F3).
The partition search strategy allows our IEO module to be used for features of the smaller receptive field, which overcomes the limitation of the global perception strategy in~\cite{2019-CVPR-AFNet}.
If we want to use the IEO for more shallow features, we can further decompose the features F$_{A}$, F$_{B}$, F$_{C}$ and F$_{D}$ (left side of Fig.\ref{IEO}) into smaller sub-regions for partition search and integration before step1 and step2.
To keep the network simple, we only used partitioning and integration once in IEO, and let IEOs be used for features i3, i4, i5 in PRNet of Fig.\ref{pipline}.
% The problem of too-large void ratio can be solved by partitioning and stacking PVMs (F1$\rightarrow$F2$\rightarrow$F3).

The disadvantage of atrous convolution with a large atrous ratio is that the information given by spatial continuity may be lost (such as edges).
Because the sampling points of atrous convolution are too scattered and the information of a single sampling point is not continuous enough.
This is the same as the human eye can't pay attention to the details of objects if there is only peripheral vision.
The human eye diagram on the left of Fig.~\ref{IEO} shows the foveal vision cells and peripheral vision cells.
Foveal visual cells are rich and concentrated, and they can observe the details of the object.
The peripheral visual cells are sparse and widely distributed, and they can organize a broad space scene~\cite{PV2}\cite{PV3}.
They are like convolutions of different atrous rates.
How to balance these two capabilities is the purpose of IEO module design.

% However, PVM (F1, F2 in Fig.\ref{IEO-detail}) has 9 sampling points.
The dark orange dots in F3 of Fig.\ref{IEO-detail} can be regarded as four groups of sampling points of PVM.
One sampling point of PVM consists of 9 points (3x3 Conv).
Therefore, PVM has both a large atrous rate and detail observation ability.
It is worth noting that the 3x3 convolution here can also use atrous convolution with a small atrous rate.
For simplicity, we use normal convolution here.
The receptive field of the dark orange pixel in F3$^{'}$ can cover the dark green areas in F1.
The receptive field of F3$^{'}$ is global.
In addition, we add the fovea vision feature (Fig.\ref{IEO} the red arrow) to cooperate with the P-feature.
In the last stage of step 3, we use concatenation operation to merge the original feature, fovea vision feature (r=5), and peripheral vision feature. Finally, we use PR block to balance the relationship among the three types of features.

\begin{scriptsize}
\begin{align}\label{map}
&\mathbf{F3^{'}} = PVM(\mathbf{F3}),&\\
& W_{1}^{'},W_{2}^{'},W_{3}^{'}  = \frac{3*e^{W_{1}}}{\sum_{j=1}^3 e^{W_{j}}}, \frac{3*e^{W_{2}}}{\sum_{j=1}^3 e^{W_{j}}}, \frac{3*e^{W_{3}}}{\sum_{j=1}^3 e^{W_{j}}},&\\
&\mathbf{F4} = Cat_{c}(  W_{1}^{'}*\mathbf{F3^{'}}, W_{2}^{'}*C^{r=5}(\mathbf{F1}), W_{2}^{'}*\mathbf{F1}   ),&
\end{align}
\end{scriptsize}
where $PVM(\cdot)$ is to repeat Eq.7-9 for $\mathbf{F3}$. $W_{1}, W_{2}, W_{3}$ are the weight produced by the memory unit.
Inspired by the classification network, Eq.13 uses a softmax layer to associate weights.
SAC uses a similar approach (1-S) to associate partial weights~\cite{2020-arxiv-detectRS}.
The three features regulated are peripheral vision feature, fovea vision feature (an original feature after atrous convolution operation), and original feature.
The atrous ratio of the convolution of the foveal vision feature is r = 5, which is designed according to the experimental effect.
% We discuss the 3 types of features and why r = 5 in supplementary materials.

\section{Global-Local Transformer}
\begin{figure}[t]% The dilation operation bring the influence of weight of edge pixels to the surrounding pixels.
\centering
\includegraphics[width=0.9\columnwidth]{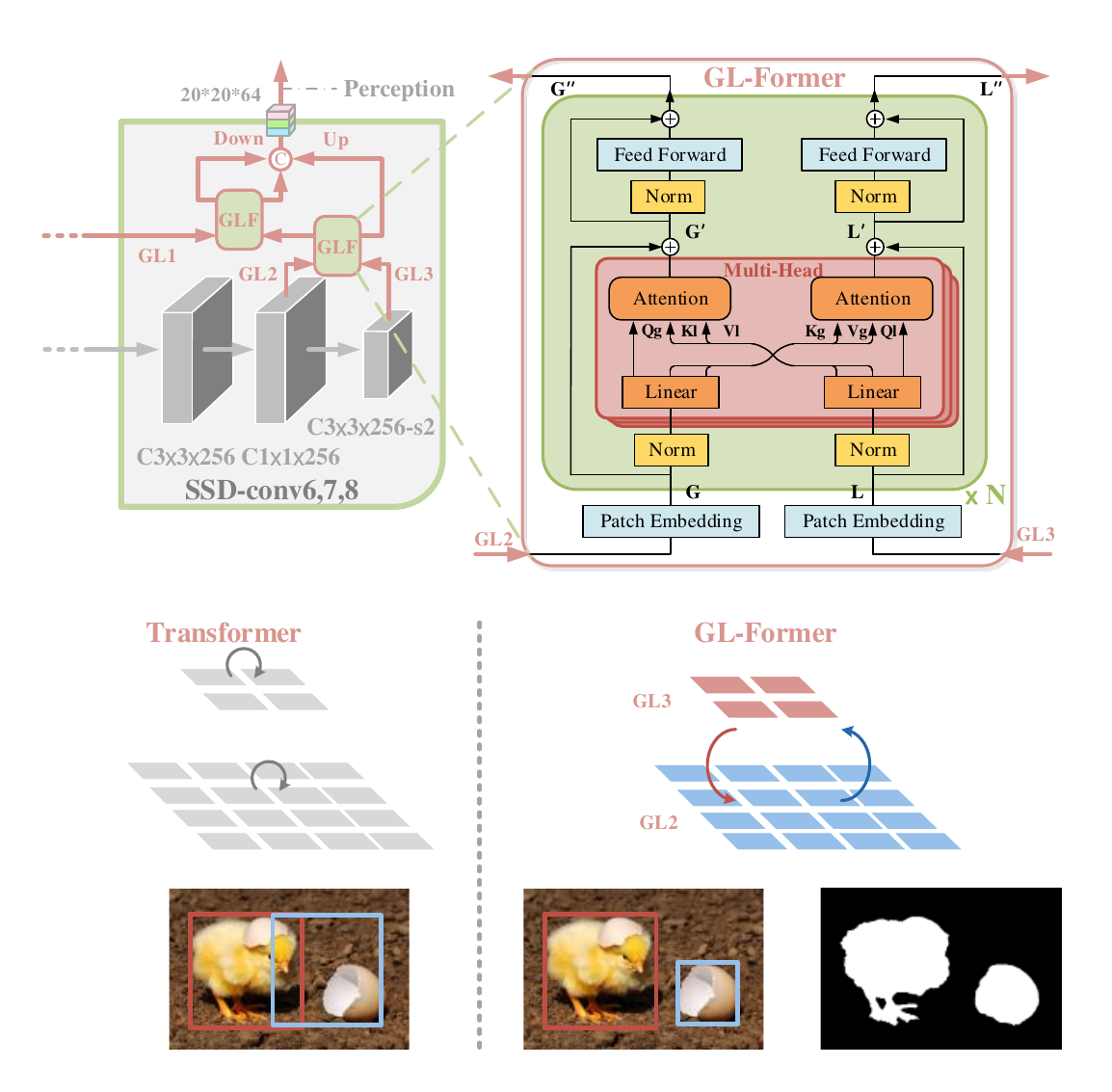}
\caption{The detailed description of global-local transformer and the application on backbone.
}
\label{GL-Former}
\end{figure}

Because the network structure adopts the GSS structure, the positioning accuracy of the high-level features is very important.
We have carried out association analysis on cross-regional features through IEO modules to optimize the semantic information of high-level features.
In this section, we propose a global-local transformer module (GL-Former) to further optimize the highest-level features.
Traditional transformer networks build long-distance dependencies on different regions of the same layer features.
The tokens of the same layer in a traditional transformer have the same receptive field range.
The tokens of different layers have different receptive fields.
The GL-Former we designed uses tokens with different receptive fields to compare and analyze objects of different sizes and better capture salient objects.
As shown at the bottom of Fig.\ref{GL-Former}, the design motivation of GL-Former is visually displayed in the form of pictures.

We add GL-Former to optimize the SSD structure of the backbone in Fig.\ref{pipline}.
The improved structure is shown in Fig.\ref{GL-Former}, the output features GL1, Gl2, and GL3 are optimized twice by two GL-Former modules.
Within GL-Former, we change the structure and input to improve the self-attention mechanism into an interactive attention mechanism.
This interactive multi-head attention mechanism can help tokens with different receptive fields interact better.
The tokens obtained from deeper features have global perception ability, and the tokens obtained from shallower features can capture local information.

\section{Supervision}
We use the widely used binary cross entropy (BCE) loss and a consistency-enhanced (CEL) loss~\cite{2020-CVPR-MINet} to supervise the prediction map, as shown in Eq.\ref{BCE-IOU}.
\begin{equation}\label{BCE-IOU}
L = L_{bce} + L_{cel}.
\end{equation}

% => DEL5
%\textcolor{blue}{BCE loss is widely used in binary classification and segmentation.
%But BCE loss has some disadvantages.
%It calculates the loss of each pixel independently without fully considering the global structure of the image.
%Besides, the loss of foreground pixels will be diluted, when the images are dominated by the background.
%So we adopt CEL loss to help the network pay more attention to the global structure in the training process.
%Besides, CEL loss is not affected by the unbalanced distribution of foreground and background, and it can strongly penalize the situation that the predicted foreground region is disjoint from the ground truth in a %large range.}

BCE loss is defined as:
\begin{scriptsize}
\begin{equation}\label{loss}
L_{bce} = - \sum_{(x,y)}[g(x,y)log(p(x,y))+(1-g(x,y))log(1-p(x,y))],
\end{equation}
\end{scriptsize}
where $p(x,y) \in [0,1]$ is the prediction result of saliency map at $(x,y)$.
$g(x,y) \in [0,1]$ is the ground truth label of the pixel $(x,y)$.

CEL loss is a variant of IoU loss, which can measure the similarity of two images from an overall perspective.
it is defined as:
\begin{scriptsize}
\begin{equation}\label{iou-b}
L_{cel}=\frac{\sum_{(x,y)}[p(x,y)+g(x,y)-2*g(x,y)*p(x,y)]}  {\sum_{(x,y)}[g(x,y)+p(x,y)] }.\\
\end{equation}
\end{scriptsize}

\section{Experiment}
\subsection{Datasets and Evaluation Metrics }
We evaluate the proposed architecture on 5 SOD datasets:
DUTS~\cite{2017-CVPR-DUTS} with 10,553 training and 5,019 test images,
DUT-OMRON~\cite{LiThe} with 5,168 images,
ECSSD~\cite{2013-CVPR-Hierarchical} with 1,000 images,
PASCAL-S~\cite{LiThe} with 850 images, % with 1,447 images (follow the data partition of ~\cite{HouPami19Dss})
HKU-IS~\cite{2015-CVPR-HKU-IS} with 4,447 images.
We follow the data partition of~\cite{2020-CVPR-MINet}\cite{2019-TPAMI-DSS} to use 1,447 images of HKU-IS for testing.

\begin{figure}[tb]% The dilation operation bring the influence of weight of edge pixels to the surrounding pixels.
\centering
\includegraphics[width=0.8\columnwidth]{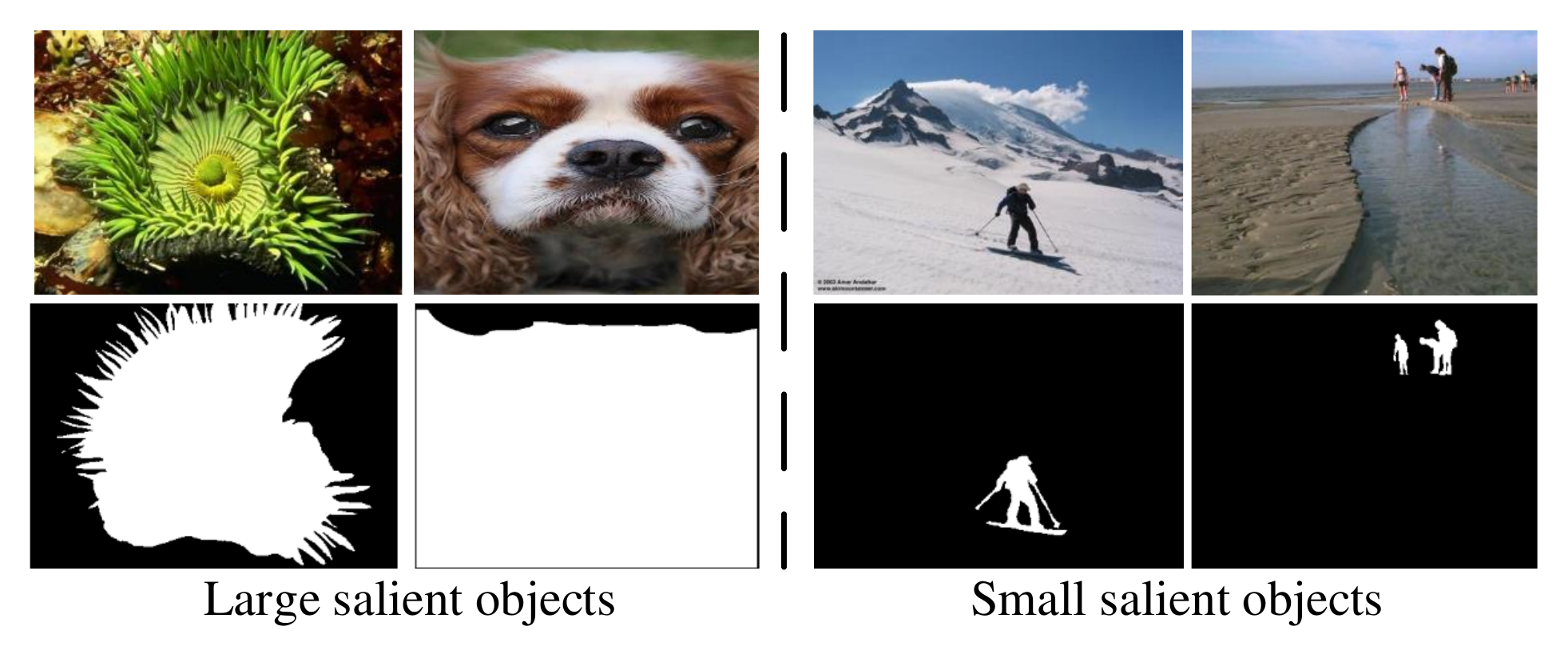}
\caption{On the left are images of a large salient object dataset. On the right are images of a small salient object dataset.}% Here $\lambda$ is 0.
\label{P-BS}
\end{figure}

In addition, we define a large salient object dataset (L) and a small salient object dataset (S).
They are helpful to further analyze the dynamic regulation of PR block when dealing with different scale objects.
We select large object images (1270) and small object images (1576) based on the ratio of white pixels in the GT, as shown in Eq.\ref{E-BS} and Fig.\ref{P-BS}.
% to the proportion of GT's white pixels to the total number of pixels.
The pictures and ground truth labels are selected from five common test datasets (ECSSD, PASCAL-S, DUT-OMRON, DUTS, HKU-IS).
$P_{w}$ is the number of white pixels in the GT and $P_{b}$ is the number of black pixels.
$t_{1}, t_{2}$ is the threshold.
We set $t_{1}$ and $t_{2}$ to 0.38 and 0.03 respectively to obtain dataset L and S.

\begin{equation}
\left\{
                 \begin{array}{lr}
                 img \in $L$,  & if \frac{P_{w}}{P_{w} + P_{b}} > $t1$  \\
                 img \in $S$, & if \frac{P_{w}}{P_{w} + P_{b}} < $t2$.
                 \end{array}
\right.
\label{E-BS}
\end{equation}

We use six metrics to evaluate the performance of PRNet and other state-of-the-art models.
% MAE F F max F avg
Mean absolute error ($\textbf{MAE}$)~\cite{perazzi2012saliency} measures the average pixel-level relative error between the prediction and the GT by calculating the mean of the absolute value of the difference.
F-measure ($\textbf{F}_{\beta}$)~\cite{2009-CVPR-Frequency} are also widely adopted in previous models~\cite{2019-ICCV-EGNet}\cite{2019-TPAMI-DSS}.
$F_{\beta}$ is the weighted harmonic mean of Precision and Recall and $\beta^{2}$ is usually set to 0.3.
The maximal $F_{\beta}$ values are calculated from PR curves, represented as $\textbf{F}_{max}$.
An adaptive threshold (twice the mean value of the prediction) is adapted to calculate $\textbf{F}_{avg}$.
%We calculate the maximal $\textbf{F}_{\beta}$ values from PR curves, denoted as $\textbf{F}_{max}$ and use an adaptive threshold that is twice the mean value of the prediction to calculate $\textbf{F}_{avg}$.
Weighted F-measure ($\textbf{F}^{w}_{\beta}$) is a measure of completeness for improving F-measure~\cite{2014-CVPR-FW}.
Structural similarity measure ($\textbf{S}_{\alpha}, \alpha = 0.5$)~\cite{2017-ICCV-Sm} and E-measure ($\textbf{E}_{m}$)~\cite{2018-IJCAI-Em} are also useful for quantitative evaluation of saliency maps.
Besides, precision-recall (\textbf{PR}) curves are drawn. % fan2017structure-measure

\subsection{Implementation}
%We combine the training datasets of CAMO-Train, CPD1K-Train, COD10K-Train and take them as the COD training dataset, which follows SINet~\cite{fan2020Camouflage}.
We follow~\cite{2020-AAAI-F3}\cite{2020-CVPR-MINet}\cite{2019-ICCV-SCRN} to use DUTS-TR~\cite{2017-CVPR-DUTS} as the training dataset and other above-mentioned datasets are used as testing datasets.
% We use DUTS-TR~\cite{Wang2017Learning} as training dataset.
In the training phase, we follow~\cite{2020-CVPR-MINet} to use random horizontal flipping, random color jittering, and random rotating as data augmentation techniques to prevent the over-fitting problem.
PRNet is trained for 40 epochs on an NVIDIA RTX 2080Ti GPU. % To ensure model convergence   with a mini-batch of 4
Batchsize is set to 4.
VGG-16, pre-trained on the ImageNet dataset, is used as the backbone network.
The parameters for the rest of PRNet are initialized by the default setting of PyTorch.
% VGG16 is adapted as backbone and its parameters are initialized with the corresponding models pre-trained on the ImageNet dataset.
Our model adopts the stochastic gradient descent (SGD) optimizer with a momentum of 0.9, a weight decay of 0.0005, and an initial learning rate of 0.001.
"Poly" strategy~\cite{2015-CS-ParseNet} (factor is 0.9) is applied.
During testing, the input size is set to 320x320.

% Tab1-FPN
\begin{table}[t]
  \centering
  \caption{Ablation experiments on various structures and modules.
  Because strategy 2 (PR$_{s}$) performs best in the final network structure (PRNet), to simplify the experiment, we use strategy PR$_{s}$ for PR block in the experiments after Line 4.}
  \setlength{\tabcolsep}{1mm}{
    \scriptsize
    \begin{tabular}{|l|rrrr|rrrr|}
    \hline
    \multicolumn{1}{|c|}{Model} & \multicolumn{4}{c|}{DUTS-TE}  & \multicolumn{4}{c|}{DUT-OMRON} \\
          & \multicolumn{1}{c}{F$^{w}_{\beta}$} & \multicolumn{1}{c}{E$_{m}$} & \multicolumn{1}{c}{S$_{m}$} & \multicolumn{1}{c|}{MAE} & \multicolumn{1}{c}{F$^{w}_{\beta}$} & \multicolumn{1}{c}{E$_{m}$} & \multicolumn{1}{c}{S$_{m}$} & \multicolumn{1}{c|}{MAE} \\
    \hline
    1.FPN   & .725  & .859  & .842  & .057  & .629  & .813  & .780  & .078 \\
    2.FPN(PR$_{FC}$) & \textcolor[rgb]{ 1,  0,  0}{.773} & \textcolor[rgb]{ 1,  0,  0}{.890} & \textcolor[rgb]{ 1,  0,  0}{.857} & \textcolor[rgb]{ 1,  0,  0}{.047} & \textcolor[rgb]{ 1,  0,  0}{.666} & \textcolor[rgb]{ 1,  0,  0}{.842} & \textcolor[rgb]{ 1,  0,  0}{.793} & \textcolor[rgb]{ 1,  0,  0}{.067} \\
    3.FPN(PR$_{C}$) & .764  & .887  & .852  & .048  & .658  & .839  & .788  & .068 \\
    4.FPN(PR$_{S}$) & .762  & .883  & .853  & .048  & .655  & .836  & .788  & .069 \\
    5.FPN(PR$_{S}$-fixed) & .735  & .866  & .840  & .053  & .617  & .810  & .769  & .078 \\
    6.FPN(GateNet)~\cite{2020-ECCV-GateNet} & .744  & .886  & .837  & .051  & .624  & .839  & .767  & .071 \\
    \hline
    7.FPN+AIMs~\cite{2020-CVPR-MINet} & .768  & .884  & .860  & .047  &- & -  & -  & - \\
    \hline
    8.FPNssd(PR$_{S}$) & .774  & .889  & .861  & .047  & .675  & .839  & .802  & .067 \\
    \hline
    9.GGS   & .751  & .880  & .845  & .051  & .641  & .828  & .777  & .073 \\
    10.GGS(PR$_{S}$) & \textcolor[rgb]{ 1,  0,  0}{.766} & \textcolor[rgb]{ 1,  0,  0}{.892} & \textcolor[rgb]{ 1,  0,  0}{.851} & \textcolor[rgb]{ 1,  0,  0}{.047} & \textcolor[rgb]{ 1,  0,  0}{.651} & \textcolor[rgb]{ 1,  0,  0}{.841} & \textcolor[rgb]{ 1,  0,  0}{.783} & \textcolor[rgb]{ 1,  0,  0}{.068} \\
    \hline
    11.FPN+CBAM~\cite{2018-ECCV-CBAM} & .768  & .888  & .856  & .048  & .661  & .837  & .792  & .069 \\
    12.FPN+CBAM(PR$_{S}$) & \textcolor[rgb]{ 1,  0,  0}{.775} & \textcolor[rgb]{ 1,  0,  0}{.894} & \textcolor[rgb]{ 1,  0,  0}{.858} & \textcolor[rgb]{ 1,  0,  0}{.046} & \textcolor[rgb]{ 1,  0,  0}{.666} & \textcolor[rgb]{ 1,  0,  0}{.847} & \textcolor[rgb]{ 1,  0,  0}{.793} & \textcolor[rgb]{ 1,  0,  0}{.066} \\
    \hline
    13.FPN+CFE~\cite{2019-CVPR-PFANet} & .793  & .896  & .867  & .044  & .688  & .849  & .805  & .062 \\
    14.FPN+CFE(PR$_{S}$) & \textcolor[rgb]{ 1,  0,  0}{.798} & \textcolor[rgb]{ 1,  0,  0}{.902} & \textcolor[rgb]{ 1,  0,  0}{.869} & \textcolor[rgb]{ 1,  0,  0}{.043} & \textcolor[rgb]{ 1,  0,  0}{.700} & \textcolor[rgb]{ 1,  0,  0}{.856} & \textcolor[rgb]{ 1,  0,  0}{.812} & \textcolor[rgb]{ 1,  0,  0}{.060} \\
    \hline
    15.EGNet~\cite{2019-ICCV-EGNet} & .802  & .897  & .878  & .044  & .728  & .864  & .836  & .057 \\
    16.EGNet(PR$_{S}$)~\cite{2019-ICCV-EGNet} & \textcolor[rgb]{ 1,  0,  0}{.806} & \textcolor[rgb]{ 1,  0,  0}{.900} & \textcolor[rgb]{ 1,  0,  0}{.881} & \textcolor[rgb]{ 1,  0,  0}{.042} & \textcolor[rgb]{ 1,  0,  0}{.731} & \textcolor[rgb]{ 1,  0,  0}{.866} & \textcolor[rgb]{ 1,  0,  0}{.837} & \textcolor[rgb]{ 1,  0,  0}{.056} \\
    \hline
    17. CFD~\cite{2020-AAAI-F3}           & .778              & .890              & .858              & .047              & .674              & .836              & .796              & .066 \\
    18. CFD(PR$_{S}$)~\cite{2020-AAAI-F3} & \textcolor[rgb]{ 1,  0,  0}{.790} & \textcolor[rgb]{ 1,  0,  0}{.891} & \textcolor[rgb]{ 1,  0,  0}{.861} & \textcolor[rgb]{ 1,  0,  0}{.044} & \textcolor[rgb]{ 1,  0,  0}{.681}   & \textcolor[rgb]{ 1,  0,  0}{.842} & \textcolor[rgb]{ 1,  0,  0}{.799} & \textcolor[rgb]{ 1,  0,  0}{.062} \\
    \hline
    \end{tabular}%
  \label{FPN}%
  }
\end{table}%

% Tab2-XR
\begin{table}[t]% GGS(PRs) & .766  & .892  & .851  & .047  & .651  & .841  & .783  & .068
  \centering
  \caption{Ablation experiment of PRNet.
  Because strategy 2 (PR$_{s}$) performs best in the final network structure (PRNet 3rd line), to simplify the experiment, PR$_{s}$ is used in every step of the ablation experiment. The model in the 6th line is the same as that in the 3rd line, but the loss only uses BCE loss (CEL loss is removed).}
  \setlength{\tabcolsep}{0.8mm}{
    \scriptsize
    \begin{tabular}{|l|rrrr|rrrr|}
    \hline
    \multicolumn{1}{|c|}{Model} & \multicolumn{4}{c|}{DUTS-TE}  & \multicolumn{4}{c|}{DUT-OMRON} \\
          & \multicolumn{1}{c}{F$^{w}_{\beta}$} & \multicolumn{1}{c}{E$_{m}$} & \multicolumn{1}{c}{S$_{m}$} & \multicolumn{1}{c|}{MAE} & \multicolumn{1}{c}{F$^{w}_{\beta}$} & \multicolumn{1}{c}{E$_{m}$} & \multicolumn{1}{c}{S$_{m}$} & \multicolumn{1}{c|}{MAE} \\
    \hline
    % Baseline & .725  & .859  & .842  & .057  & .629  & .813  & .780  & .078\\
    1. GGS$_{ssd}$(PR$_{S}$) & .755  & .877  & .859  & .050  & .660  & .837  & .801  & .069 \\ % & .766  & .892  & .851  & .047  & .651  & .841  & .783  & .068
    2. GGS$_{ssd}$(PR$_{S}$) + IEO & .793  & .901  & .868  & .043  & .692  & .851  & .809  & .062 \\
    3. GGS$_{ssd}$(PR$_{S}$) + IEO(PR$_{S}$) & .802 & .908 & .872 & .041 & .698 & .857 & .812 & .059 \\
    4. GGS$_{GLF}$(PR$_{S}$) + IEO(PR$_{S}$) & \textcolor[rgb]{ 1,  0,  0}{.803} & \textcolor[rgb]{ 1,  0,  0}{.908} & \textcolor[rgb]{ 1,  0,  0}{.875} & \textcolor[rgb]{ 1,  0,  0}{.041} & \textcolor[rgb]{ 1,  0,  0}{.700} & \textcolor[rgb]{ 1,  0,  0}{.859} & \textcolor[rgb]{ 1,  0,  0}{.814} & \textcolor[rgb]{ 1,  0,  0}{.059} \\
    \hline
    \hline
    5. GGS(PR$_{S}$)     + IEO(PR$_{S}$) & .794  & .903  & .868  & .042  & .690  & .856  & .807  & .060 \\
    6. GGS$_{ssd}$(PR$_{S}$) + CFE(PR$_{S}$)  & .794	& .902	& .869	& .043  &.696	&.856	&.812	& .061\\
    7. w/o CEL &  .765 & .879 & .865 & .048 & .669 & .840 & .808 & .067 \\

    %GGS$_{ssd}$(PR$_{FC}$) + IEO(PR$_{FC}$) & .794  & .904  & .868  & .042  & .690  & .852  & .807  & .062 \\
    %GGS$_{ssd}$(PR$_{C}$) + IEO(PR$_{C}$) & .792  & .902  & .868  & .043  & .686  & .851  & .806  & .062 \\
    \hline
    \end{tabular}%
    }
  \label{xiaorong}% & .766  & .892  & .851  & .047  & .651  & .841  & .783  & .068 \\
\end{table}%

% Tab3-IEO
\begin{table}[t]
  \small
  \centering
  \caption{Ablation Experiment of IEO Module.
  We adjust the location and number of IEO on the basis of PRNet(PR$_{S}$).
  The experiments show that the best location is i3 i4 i5.}
  \setlength{\tabcolsep}{1mm}{
    \scriptsize
    \begin{tabular}{|l|rrrr|rrrr|}
    \hline
    \multicolumn{1}{|c|}{Model} & \multicolumn{4}{c|}{DUTS-TE}  & \multicolumn{4}{c|}{DUT-OMRON} \\
          & \multicolumn{1}{c}{F$^{w}_{\beta}$} & \multicolumn{1}{c}{E$_{m}$} & \multicolumn{1}{c}{S$_{m}$} & \multicolumn{1}{c|}{MAE} & \multicolumn{1}{c}{F$^{w}_{\beta}$} & \multicolumn{1}{c}{E$_{m}$} & \multicolumn{1}{c}{S$_{m}$} & \multicolumn{1}{c|}{MAE} \\
    \hline
    1. IEO in i5 & .790  & .901  & .867  & .043 & .684	&.849	& .805	& .063\\
    2. IEO in i4, i5 & .792  & .905  & .867  & .042 & .682	& .850	& .803	& .063\\
    3. IEO in i3, i4, i5 & \textcolor[rgb]{ 1,  0,  0}{.802} & \textcolor[rgb]{ 1,  0,  0}{.908} & \textcolor[rgb]{ 1,  0,  0}{.872} & \textcolor[rgb]{ 1,  0,  0}{.041} &.698 & .857 & .812 & \textcolor[rgb]{ 1,  0,  0}{.059} \\
    4. IEO in i2, i3, i4, i5 & .797  & .904  & .870  & .042     & \textcolor[rgb]{ 1,  0,  0}{.699}	& \textcolor[rgb]{ 1,  0,  0}{.859}	& \textcolor[rgb]{ 1,  0,  0}{.813}	& .060\\
    5. IEO in i1, i2, i3, i4, i5 & .798  & .903  & .871  & .042     & \textcolor[rgb]{ 1,  0,  0}{.699}	& .856	& \textcolor[rgb]{ 1,  0,  0}{.813}	& .061\\
    \hline
    \end{tabular}%
    }
  \label{T-IEO}%
\end{table}%

% Tab4-FCS
\begin{table}[t]
  \small
  \centering
  \caption{Comparison of 3 types of perception strategies in PRNet.}
  \setlength{\tabcolsep}{1.4mm}{
    \scriptsize
    \begin{tabular}{|l|rrrr|rrrr|}
    \hline
    \multicolumn{1}{|c|}{Model} & \multicolumn{4}{c|}{DUTS-TE}  & \multicolumn{4}{c|}{DUT-OMRON} \\
          & \multicolumn{1}{c}{F$^{w}_{\beta}$} & \multicolumn{1}{c}{E$_{m}$} & \multicolumn{1}{c}{S$_{m}$} & \multicolumn{1}{c|}{MAE} & \multicolumn{1}{c}{F$^{w}_{\beta}$} & \multicolumn{1}{c}{E$_{m}$} & \multicolumn{1}{c}{S$_{m}$} & \multicolumn{1}{c|}{MAE} \\
    \hline
    1. PRNet(PR$_{FC}$) & .794  & .904  & .868  & .042  & .690  & .852  & .807  & .062 \\
    2. PRNet(PR$_{S}$) & \textcolor[rgb]{ 1,  0,  0}{.802} & \textcolor[rgb]{ 1,  0,  0}{.908} & \textcolor[rgb]{ 1,  0,  0}{.872} & \textcolor[rgb]{ 1,  0,  0}{.041} & \textcolor[rgb]{ 1,  0,  0}{.698} & \textcolor[rgb]{ 1,  0,  0}{.857} & \textcolor[rgb]{ 1,  0,  0}{.812} & \textcolor[rgb]{ 1,  0,  0}{.059} \\
    3. PRNet(PR$_{C}$) & .792  & .902  & .868  & .043  & .686  & .851  & .806  & .062 \\
    \hline
    \end{tabular}%
    }
  \label{FCS}%
\end{table}%

\subsection{Ablation Studies}
% => DEL6
% \textcolor{blue}{First, we analyze the performance of PR block in different network structures and different modules in Tab.\ref{FPN}.
%Then, ablation experiments for PRNet are presented in Tab.\ref{xiaorong}.
%Detailed experiments on the IEO module are also presented in Tab.\ref{T-IEO}.
%Finally, the analysis of the 3 perception strategies under simple (FPN) and complex (PRNet) structures is presented in Tab.\ref{FPN} and Tab.\ref{FCS}, respectively.}

\textbf{Ablation analysis of PR block in different structures and modules.}
Lines 1-4 of Tab.\ref{FPN} analyze the effect of PR block with different perception strategies in FPN structure.
$PR_{FC}, PR_{C}, PR_{S}$ are strategy 1,2,3 in Fig.\ref{PRblock}.
The $PR_{FC}$ has the best regulation effect in FPN structure because of its rich parameters and intensive interaction analysis in the fully-connected layer.
But in the final network structure (PRNet), $PR_{S}$ is the best (Tab.\ref{FCS}), we will explain this phenomenon in later analysis.
In order to simplify the experiment, we uniformly use the best strategy $PR_{S}$ in the final network to carry out the comparison experiments and the ablation experiment in Tab.\ref{FPN}, Tab.\ref{xiaorong} and Tab.\ref{T-IEO}.
Line 6 of Tab.\ref{FPN} is the gate strategy provided by the GateNet~\cite{2020-ECCV-GateNet}, and its effect is not as good as the PR block of global perception and regulation.
It is worth noting that line 7 of Tab.\ref{FPN} is the AIMs module of MINet~\cite{2020-CVPR-MINet}.
AIM has a more complex interaction structure than PR block, but the regulation effect of PR block is better.
Line 8 verify the effect of the multi-level perception strategy in the FPN structure.
FPN$_{ssd}$(PR$_{S}$) means that FPN-PR uses the encoder structure shown in the left side of Fig.~\ref{pipline}.
Line 9-18 verify the improvement of PR block to GGS structure (the lower right corner of Fig.\ref{pipline}), CFE module (Fig.\ref{related} (e)), CBAM module (Fig.\ref{related} (f)), EGNet network (Fig.\ref{related} (g)) and CFD decoder (Fig.\ref{related} (b)).

\textbf{Ablation analysis of the PRNet.}
Tab.\ref{xiaorong} shows the ablation experiment of PRNet (Fig.\ref{pipline}).
The baseline is GGS$ssd$(PR$_{S}$).
IEO improves the network performance greatly in line 2, but the allocation of foveal vision feature and peripheral vision feature is not balanced.
The performance of IEO is further improved after being regulated by PR block (line 3).
The right side of Fig.~\ref{IEO} shows how the weights obtained by the perception part regulate the features of IEO.
Lines 3 and 4 show the effect of multi-level perception.
The experiment in the 5th line replaces the IEO-PR module in the 3rd line with the CFE-PR module (Fig.\ref{related} (e)), which proves the effectiveness of the IEO-PR module.
The model of the experiment in the 6th line is the same as that in the 3rd line, but the experiment in the 6th line is only supervised by BCE loss (CEL loss is removed). The 3rd and 6th experiments show the effect of CEL loss.

It is worth noting that, as shown in Fig.\ref{pipline}, the IEO and CFE modules in Tab.\ref{xiaorong} are placed in i3, i4, and i5 positions, which follows the setting of ~\cite{2019-CVPR-PFANet}.
The peripheral vision module in the IEO module has a large void rate, so it is better to use it for high-level features with a larger receptive field.
Experiments in Tab.\ref{T-IEO} also show that this scheme is the best for IEO-PR module.
Besides, reducing the number of IEO modules is conducive to simplify the model and improving the speed.

\textbf{Perception strategy analysis.}
Lines 1,2,3,4 of Tab.\ref{FPN} show that strategy 1 (PR$_{FC}$) is the best in simple structures (FPN).
While strategy 2 (PR$_{S}$) performs best in complex structures (PRNet) as shown in Tab.\ref{FCS}.

\begin{figure}[t]
\centering
\includegraphics[width=1\columnwidth]{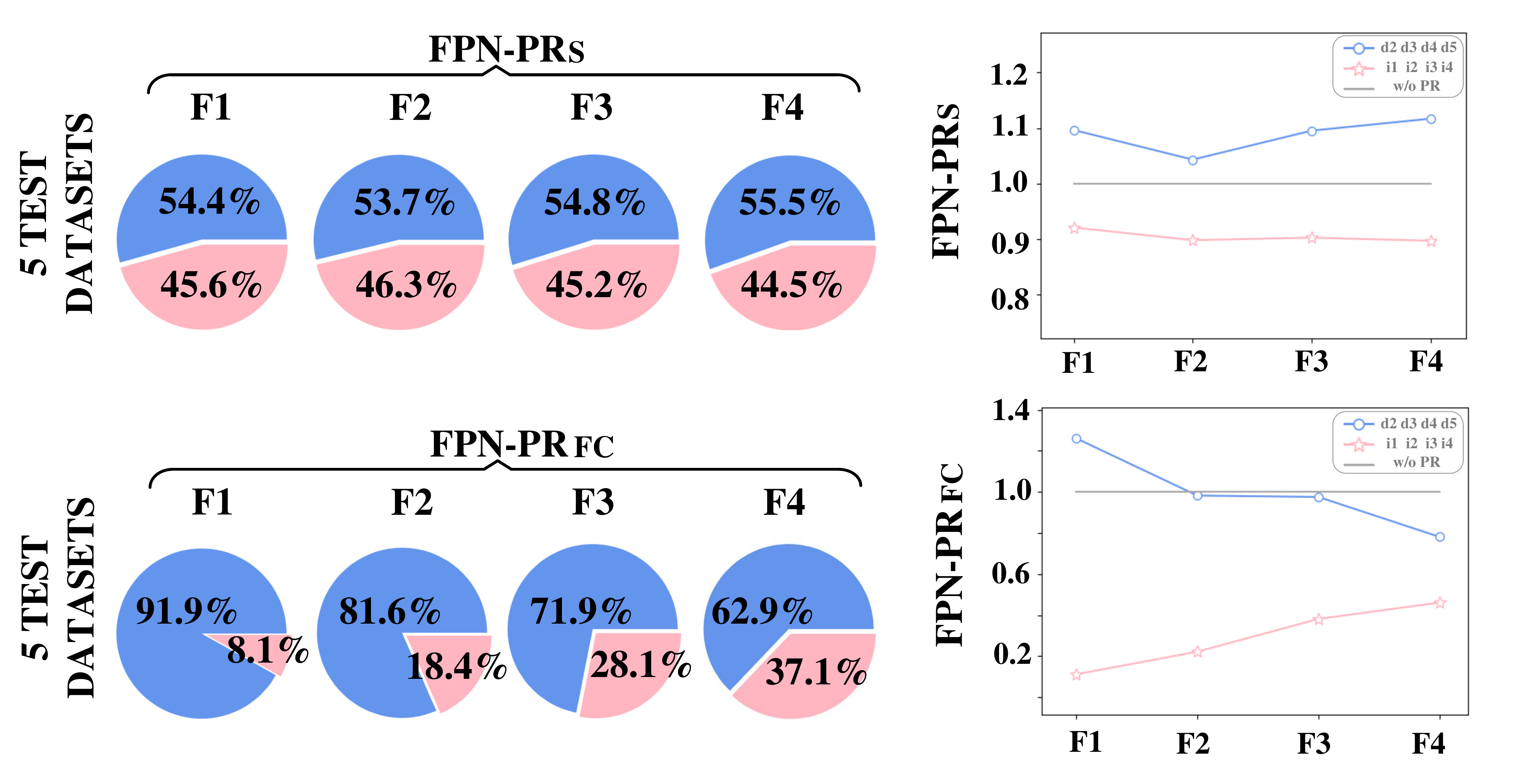}
\caption{The effect analysis of strategy 1 (PR$_{FC}$) and strategy 2 (PR$_{S}$) in FPN architecture.
The pie chart and line chart show the statistical results of feature weights on five test datasets.
The pie chart is presented in the proportion of feature weight.
The line chart is presented in the form of concrete numerical value.
F1-F5 represent the five feature fusion processes of FPN.
F1 represents the feature fusion process of i1 and d2.
The blue regions represent the weight ratio of decoder features (d2, d3, d4, d5).
The pink regions represent the weight ratio of interlayer features (i1, i2, i3, i4).
For the feature definitions of i1 and d2, please refer to FPN-PR in the upper right corner of Fig.\ref{pipline}.}
\label{FC-S}
\end{figure}

The fully-connected layers in PR$_{FC}$ that are inspired by the classification network make the weights coupled and correlation strong as shown in Fig.\ref{PRblock}.
In addition, there are many parameters in the FC layer, which is also helpful to recalibrate the weights of features with obvious differences in the simple structure (FPN). But for complex network PRNet, the difference between the features to be fused becomes smaller.
Take the feature fusion process F$_{G}$ on the far right of PRNet (Fig.\ref{pipline}) as an example, because d4 is the fusion feature of g2 (d5) and i4, the difference between g2 and d4 is reduced.
Fully-connected layers in strategy 1 over-interpret the weights, which makes the effect of PR$_{FC}$ block worse.
Fully-connected layers are also used in strategy 3 (PR$_{C}$), so it has the same problem.
% Strategy 3 (PR$_{C}$) is designed to imitate SENet.
The MU of strategy 3 evaluates each feature weight independently, which is different from the strong coupling in strategy 1.
Strategy 3 can be considered as a simplified version of strategy 1 because the parameters of its fully-connected layer are less than strategy 1.
% To be concise, we only add two additional comparative experiments (EGNet(PR$_{S})$, EGNet(PR$_{FC}$), CFD($_{S}$), CFD($_{FC}$)) for typical strategies 1 and 2, as shown in Tab.x of appendix.
Strategy 1 performs better on simple networks (FPN, CFD), and strategy 2 performs better on complex networks (PRNet, EGNet).
Besides, we analyze the weights of strategies 1 and 2 in the FPN structure, as shown in Fig.\ref{FC-S}.
We can find that strategy 1 is radical and sensitive, while strategy 2 is conservative and restricted.
These characteristics result in the performance differences between the two strategies in simple network structure and complex network structure.

PR$_{S}$ directly uses global average pooling to reduce the spatial dimension (H, W) to (1, 1), which is beneficial for complex networks (PRNet) with little difference in features to be fused.
Because there is no fully-connected layer, the final weights are more directly and closely related to the spatial features of salient objects.
PR$_{S}$ can prevent overfitting of the fully-connected layers in strategy 1, 3.

PR block is originally inspired by the classification network.
And we find that the paper, Network In Network (NIN)~\cite{2014-ICLR-NIN}, just verifies this phenomenon from the perspective of the classification task.
NIN explains the advantages of global average pooling to replace the fully connected layer in detail.
According to the above analysis, we can use strategy 1 to regulate the feature fusion process with large feature differences, while strategy 2 can be used for the feature fusion process with small feature differences.
In addition, it should be noted that strategy 1 is not suitable for all feature fusion processes.
For the case that the difference between the features to be fused is very small, radical weight regulation may have a negative impact.

\begin{figure}[htb]% The dilation operation bring the influence of weight of edge pixels to the surrounding pixels.
\centering
\includegraphics[width=0.95\columnwidth]{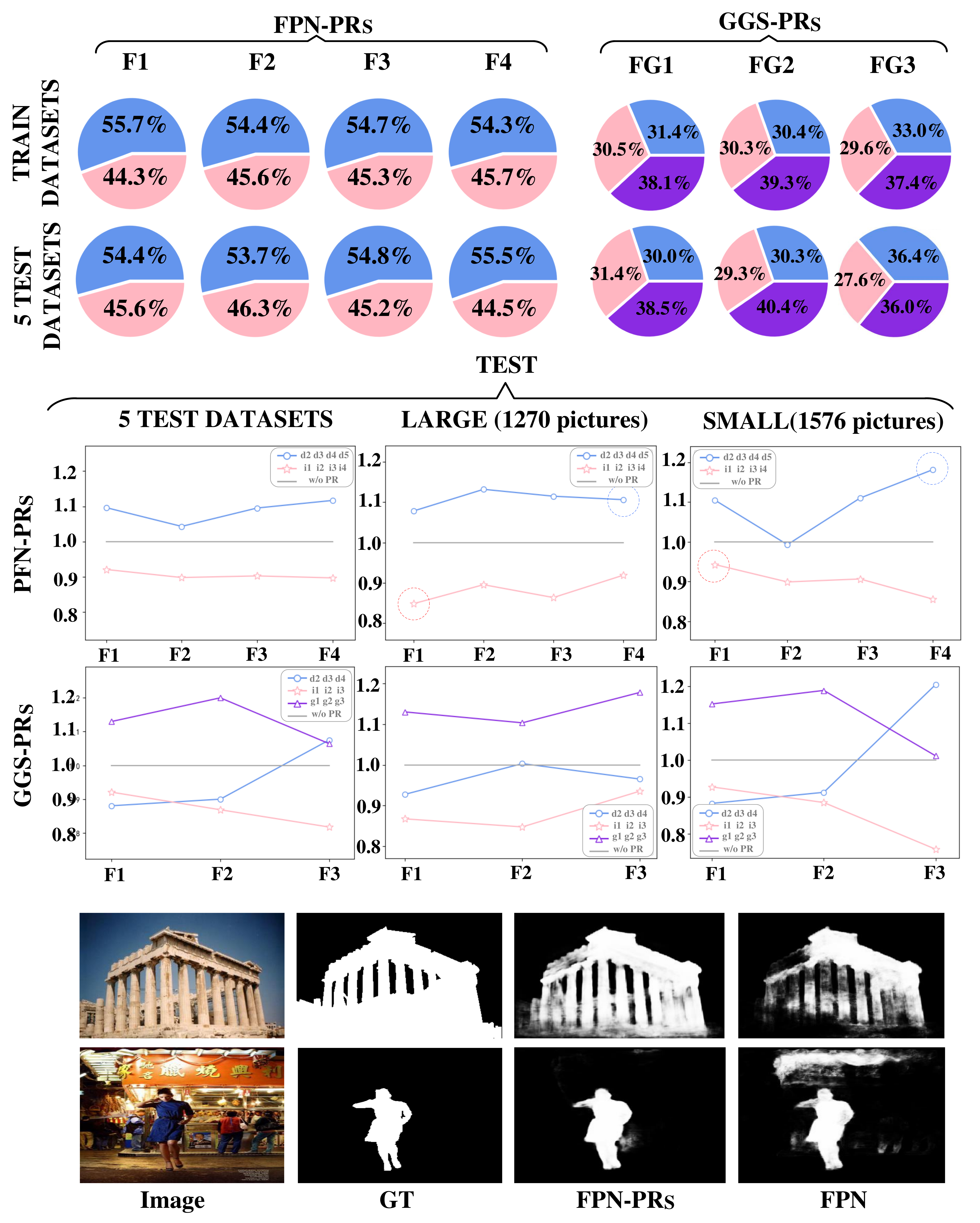} % Reduce the figure size so that it is slightly narrower than the column. Don't use precise values for figure width.This setup will avoid overfull boxes.
\caption{Weight analysis.
For FPN-PR$_{S}$ and GGS-PR$_{S}$,
we use pie charts and line charts to show the average values of multiple fusion position weights of 5 test datasets and the training dataset.
In the line chart, we use 1270 images of large objects and 1576 images of small objects respectively to show how the PR block regulates the network in the segmentation task of objects of different scales.
}% Here $\lambda$ is 0.
\label{pie}
\end{figure}
\textbf{Feature weight analysis.} To further analyze how the PR block works, we show the average weight of each fusion position in the pie chart and line chart in Fig.\ref{pie}.
F1-4 and FG1-3 represent multiple regulated points of FPN-PR and GGS-PR (Fig.\ref{pipline}).
The pie chart shows the results in the training dataset and 5 test datasets.
The line chart shows the results of 5 test datasets and the results of 1300 pictures with large objects and 1300 pictures with small objects.
The images of large objects and small objects are obtained by setting threshold according to the area ratio of white pixels in GT.
Blue, red and purple regions and lines represent decoder features, interlayer features, and global features respectively.
The grey line indicates that the weight of the feature without PR block regulation is 1.

We use the dotted circle to indicate the position where the weight of the PR block changes obviously.
This change is strongly related to the size of the salient object.
It is worth noting that the highest-level features (d5) of small objects are specially enhanced (dotted circle in the small object) to prevent dilution.
The lowest-level (i1) features in large objects are sufficiently suppressed to prevent interference.
The left side of Fig.\ref{pie} shows the effect of the PR block.
In order to verify that the dynamic regulation of weights is meaningful, we lock the weights in Tab.\ref{FPN} line 5.
The weights of FPN(PR$_{S-fixed}$) is obtained from the average weight of training dataset.
Line
\begin{table*}[htbp]
  \centering
  \caption{Quantitative evaluation. We compare 22 SOD methods on 5 SOD datasets. The maximum and mean F-measure (larger is better), S-measure (larger is better) and MAE (smaller is better) of different salient object detection methods on five benchmark datasets. The best three results are highlighted in $\textcolor[rgb]{ 1,  0,  0}{red}, \textcolor[rgb]{ .357,  .608,  .835}{blue}$ and $\textcolor[rgb]{ .439,  .678,  .278}{green}$. -V:
VGG16 as backbone in the algorithm that provides multiple backbones. *: We compare the PoolNet algorithm that uses only the DUTS-TE dataset for training. -: The author do not provide the test results of the corresponding datasets.
}
  \resizebox{\textwidth}{!}{
    \begin{tabular}{|l|cccccc|cccccc|cccccc|cccccc|cccccc|}
    \hline
    \multicolumn{1}{|c|}{Model} & \multicolumn{6}{c|}{DUTS-TE}          & \multicolumn{6}{c|}{DUT-OMRON}                        & \multicolumn{6}{c|}{HKU-IS}                   & \multicolumn{6}{c|}{ECSSD}                    & \multicolumn{6}{c|}{PASCAL-S} \\
          & F$_{max}$  & F$_{avg}$  & F$^{w}_{\beta}$     & E$_{m}$    & S$_{m}$    & MAE   & F$_{max}$  & F$_{avg}$  & F$^{w}_{\beta}$     & E$_{m}$    & S$_{m}$    & MAE   & F$_{max}$  & F$_{avg}$  & F$^{w}_{\beta}$     & E$_{m}$    & S$_{m}$    & MAE   & F$_{max}$  & F$_{avg}$  & F$^{w}_{\beta}$     & E$_{m}$   & S$_{m}$    & MAE   & F$_{max}$  & F$_{avg}$  & F$^{w}_{\beta}$     & E$_{m}$    & S$_{m}$    & MAE \\
    \hline
    DCL$_{16}$                & .782              & .712              & .408              & .839              & .734              & .088              & .757              & .695              & .583              & .830              & .772              & .080              & .757              & .695              & .583              & .830              & .772              & .080              & .901              & .874              & .790              & .908              & .870              & .068              & .830              & .776              & .685              & .836              & .794              & .108 \\
    Amulet$_{17}$             & .778              & .671              & .655              & .798              & .803              & .085              & .743              & .647              & .625              & .784              & .780              & .098              & .897              & .842              & .816              & .914              & .884              & .051              & .915              & .869              & .841              & .912              & .893              & .059              & .841              & .771              & .741              & .831              & .821              & .098 \\
    NLDF$_{17}$               & .813              & .739              & .710              & .855              & .816              & .065              & .753              & .684              & .634              & .817              & .770              & .080              & .902              & .872              & .839              & .929              & .878              & .048              & .905              & .878              & .839              & .912              & .875              & .063              & .833              & .782              & .742              & .842              & .804              & .099 \\
    UCF$_{17}$               & .771              & .624              & .586              & .766              & .777              & .117              & .735              & .613              & .564              & .763              & .758              & .132              & .888              & .810              & .753              & .893              & .866              & .074              & .911              & .840              & .789              & .888              & .883              & .078              & .830              & .708              & .681              & .787              & .803              & .126 \\
    MSRNet$_{17}$             & .829              & .703              & .720              & .840              & .839              & .061              & .782              & .676              & .669              & .820              & .808              & .073              & .914              & .857              & .853              & .935              & .902              & .040              & .911              & .839              & .849              & .905              & .895              & .054              & .858              & .747              & .769              & .837              & .841              & .081 \\
    DSS$_{17}$            & .826              & .789              & .755              & .885              & .824              & .056              & .772              & .729              & .691              & .846              & .788              & .066              & .910              & \textcolor[rgb]{ .357,  .608,  .835}{.894} & .864              & .938              & .879              & .041              & .916              & .900              & .871              & .924              & .882              & .053              & .839              & .807              & .760              & .851              & .797              & .096 \\
    BMPM$_{18}$              & .852              & .745              & .761              & .863              & .862              & .049              & .774              & .692              & .681              & .839              & .809              & .064              & .920              & .871              & .860              & .938              & .907              & .039              & .928              & .868              & .871              & .916              & .911              & .045              & .864              & .771              & .785              & .847              & \textcolor[rgb]{ .439,  .678,  .278}{.845} & .075 \\
    RAS$_{18}$                & .831              & .751              & .740              & .864              & .839              & .059              & .787              & .713              & .695              & .849              & .814              & .062              & .913              & .871              & .843              & .931              & .887              & .045              & .921              & .889              & .857              & .922              & .893              & .056              & .838              & .787              & .738              & .837              & .795              & .104 \\
    PAGRN$_{18}$              & .854              & .784              & .724              & .883              & .838              & .056              & .771              & .711              & .622              & .843              & .775              & .071              & .919              & .887              & .823              & .941              & .889              & .047              & .927              & .894              & .834              & .917              & .889              & .061              & .858              & .808              & .738              & .854              & .817              & .093 \\
    C2S$_{18}$                & .811              & .717              & .717              & .847              & .831              & .062              & .759              & .682              & .663              & .828              & .799              & .072              & .898              & .851              & .834              & .928              & .886              & .047              & .911              & .865              & .854              & .915              & .896              & .053              & .857              & .775              & .777              & .850              & .840              & .080 \\
    PoolNet$_{19}^{*}$            & .876              & -                 & -                 & -                 & .043              & .817              & \textcolor[rgb]{ 1,  0,  0}{.817} & -                 & -                 & -                 & -                 & \textcolor[rgb]{ .357,  .608,  .835}{.058} & \textcolor[rgb]{ .357,  .608,  .835}{.928} & -                 & -                 & -                 & -                 & \textcolor[rgb]{ .357,  .608,  .835}{.035} & \textcolor[rgb]{ .439,  .678,  .278}{.936} & -                 & -                 & -                 & -                 & .047              & .857              & -                 & -                 & -                 & -                 & .078 \\
    AFNet$_{19}$              & .863              & .793              & .785              & \textcolor[rgb]{ .439,  .678,  .278}{.895} & .867              & .046              & .797              & \textcolor[rgb]{ .439,  .678,  .278}{.739} & .717              & \textcolor[rgb]{ .357,  .608,  .835}{.859} & .826              & .057              & .925              & .889              & .872              & \textcolor[rgb]{ .357,  .608,  .835}{.949} & .906              & \textcolor[rgb]{ .439,  .678,  .278}{.036} & .935              & .908              & .886              & \textcolor[rgb]{ .439,  .678,  .278}{.941} & \textcolor[rgb]{ .439,  .678,  .278}{.913} & \textcolor[rgb]{ .439,  .678,  .278}{.042} & \textcolor[rgb]{ .439,  .678,  .278}{.871} & \textcolor[rgb]{ .439,  .678,  .278}{.828} & .804              & \textcolor[rgb]{ .357,  .608,  .835}{.887} & .850              & .071 \\
    MLMSNet$_{19}$            & .852              & .745              & .761              & .863              & .862              & .049              & .774              & .692              & .681              & .839              & .809              & .064              & .920              & .871              & .860              & .938              & \textcolor[rgb]{ .439,  .678,  .278}{.907} & .039              & .928              & .868              & .871              & .916              & .911              & .045              & .864              & .771              & .785              & .847              & .845              & .075 \\
    PAGE$_{19}$               & .838              & .777              & .769              & .886              & .854              & .052              & .792              & .736              & \textcolor[rgb]{ .357,  .608,  .835}{.722} & \textcolor[rgb]{ 1,  0,  0}{.860} & .825              & .062              & .920              & .884              & .868              & \textcolor[rgb]{ .439,  .678,  .278}{.948} & .904              & \textcolor[rgb]{ .439,  .678,  .278}{.036} & .931              & .906              & .886              & \textcolor[rgb]{ .357,  .608,  .835}{.943} & .912              & \textcolor[rgb]{ .439,  .678,  .278}{.042} & .859              & .817              & .792              & .879              & .840              & .078 \\
    BANet-V$_{19}$            & .852              & .789              & .781              & .891              & .861              & .046              & .793              & .731              & \textcolor[rgb]{ .439,  .678,  .278}{.719} & .856              & .823              & .061              & .920              & .887              & .871              & .948              & .903              & \textcolor[rgb]{ .439,  .678,  .278}{.036} & .935              & \textcolor[rgb]{ .439,  .678,  .278}{.910} & \textcolor[rgb]{ .439,  .678,  .278}{.890} & \textcolor[rgb]{ 1,  0,  0}{.944} & \textcolor[rgb]{ .439,  .678,  .278}{.913} & \textcolor[rgb]{ .357,  .608,  .835}{.041} & .867              & .826              & .799              & .879              & .841              & .078 \\
    HRS$_{19}$                & .843              & .793              & .746              & .889              & .829              & .051              & .762              & .708              & .645              & .842              & .772              & .066              & .913              & .892              & .854              & .938              & .883              & .042              & .920              & .902              & .859              & .923              & .883              & .054              & .852              & .809              & .748              & .850              & .801              & .090 \\
    GATE-V$_{20}$             & \textcolor[rgb]{ .439,  .678,  .278}{.870} & .783              & .786              & .888              & \textcolor[rgb]{ .439,  .678,  .278}{.870} & \textcolor[rgb]{ .439,  .678,  .278}{.045} & .794              & .724              & .704              & .854              & .821              & .061              & \textcolor[rgb]{ .357,  .608,  .835}{.928} & .889              & .872              & \textcolor[rgb]{ .439,  .678,  .278}{.948} & \textcolor[rgb]{ .357,  .608,  .835}{.909} & \textcolor[rgb]{ .357,  .608,  .835}{.035} & \textcolor[rgb]{ 1,  0,  0}{.941} & .896              & .886              & .931              & \textcolor[rgb]{ 1,  0,  0}{.917} & \textcolor[rgb]{ .357,  .608,  .835}{.041} & \textcolor[rgb]{ .357,  .608,  .835}{.882} & .810              & \textcolor[rgb]{ .439,  .678,  .278}{.807} & .870              & \textcolor[rgb]{ .439,  .678,  .278}{.856} & \textcolor[rgb]{ .439,  .678,  .278}{.070} \\
    ITSD-V$_{20}$             & \textcolor[rgb]{ 1,  0,  0}{.877} & \textcolor[rgb]{ .439,  .678,  .278}{.798} & \textcolor[rgb]{ 1,  0,  0}{.814} & .893              & \textcolor[rgb]{ 1,  0,  0}{.877} & \textcolor[rgb]{ .357,  .608,  .835}{.042} & \textcolor[rgb]{ .357,  .608,  .835}{.807} & \textcolor[rgb]{ 1,  0,  0}{.745} & \textcolor[rgb]{ 1,  0,  0}{.734} & \textcolor[rgb]{ .439,  .678,  .278}{.858} & \textcolor[rgb]{ .439,  .678,  .278}{.829} & .063              & \textcolor[rgb]{ .439,  .678,  .278}{.926} & \textcolor[rgb]{ .439,  .678,  .278}{.891} & \textcolor[rgb]{ .439,  .678,  .278}{.882} & .947              & \textcolor[rgb]{ .439,  .678,  .278}{.907} & \textcolor[rgb]{ .357,  .608,  .835}{.035} & \textcolor[rgb]{ .357,  .608,  .835}{.939} & .875              & \textcolor[rgb]{ 1,  0,  0}{.897} & .918              & \textcolor[rgb]{ .357,  .608,  .835}{.914} & \textcolor[rgb]{ 1,  0,  0}{.040} & .884              & .787              & \textcolor[rgb]{ 1,  0,  0}{.824} & .857              & \textcolor[rgb]{ 1,  0,  0}{.858} & \textcolor[rgb]{ .357,  .608,  .835}{.068} \\
    FCNet$_{20}$             & .829              & .795              & .757              & .887              & .822              & \textcolor[rgb]{ .439,  .678,  .278}{.045} & .717              & .676              & .618              & .795              & .745              & .066              & -                 & -                 & -                 & -                 & -                 & -                 & -                 & -                 & -                 & -                 & -                 & -                 & .857              & \textcolor[rgb]{ .357,  .608,  .835}{.830} & .802              & \textcolor[rgb]{ .439,  .678,  .278}{.882} & .830              & \textcolor[rgb]{ .357,  .608,  .835}{.068} \\
    HVPNet$_{20}$            & .840              & .749              & .730              & .863              & .849              & .058              & \textcolor[rgb]{ .439,  .678,  .278}{.804} & .721              & .700              & .847              & \textcolor[rgb]{ 1,  0,  0}{.831} & .065              & .916              & .871              & .840              & .936              & .899              & .045              & .928              & .889              & .855              & .924              & .903              & .052              & .849              & .794              & .753              & .850              & .827              & .091 \\
    CAGNet-V$_{20}$          & .851              & \textcolor[rgb]{ 1,  0,  0}{.820} & \textcolor[rgb]{ .439,  .678,  .278}{.797} & \textcolor[rgb]{ .357,  .608,  .835}{.900} & .852              & \textcolor[rgb]{ .439,  .678,  .278}{.045} & .782              & \textcolor[rgb]{ .357,  .608,  .835}{.743} & .718              & \textcolor[rgb]{ .439,  .678,  .278}{.858} & .807              & \textcolor[rgb]{ 1,  0,  0}{.057} & .922              & .906              & \textcolor[rgb]{ 1,  0,  0}{.888} & \textcolor[rgb]{ .439,  .678,  .278}{.948} & .899              & .033              & .930              & \textcolor[rgb]{ .357,  .608,  .835}{.911} & \textcolor[rgb]{ .357,  .608,  .835}{.892} & .932              & .897              & \textcolor[rgb]{ .439,  .678,  .278}{.042} & .860              & \textcolor[rgb]{ .439,  .678,  .278}{.828} & .799              & .874              & .825              & .077 \\
    SAMNet$_{21}$            & .836              & .745              & .729              & .864              & .849              & .058              & .803              & .717              & .699              & .847              & \textcolor[rgb]{ .357,  .608,  .835}{.830} & .065              & .915              & .870              & .837              & .938              & .898              & .045              & .928              & .891              & .858              & .930              & .907              & .050              & .850              & .790              & .747              & .849              & .824              & .093 \\
    \hline
    PRNet             & \textcolor[rgb]{ 1,  0,  0}{.877} & \textcolor[rgb]{ .357,  .608,  .835}{.815} & \textcolor[rgb]{ .357,  .608,  .835}{.802} & \textcolor[rgb]{ 1,  0,  0}{.908} & \textcolor[rgb]{ .357,  .608,  .835}{.872} & \textcolor[rgb]{ 1,  0,  0}{.041} & .789              & .731              & .698              & .857              & .812              & \textcolor[rgb]{ .439,  .678,  .278}{.059} & \textcolor[rgb]{ 1,  0,  0}{.930} & \textcolor[rgb]{ 1,  0,  0}{.906} & \textcolor[rgb]{ .357,  .608,  .835}{.885} & \textcolor[rgb]{ 1,  0,  0}{.956} & \textcolor[rgb]{ 1,  0,  0}{.910} & \textcolor[rgb]{ 1,  0,  0}{.033} & \textcolor[rgb]{ .439,  .678,  .278}{.936} & \textcolor[rgb]{ 1,  0,  0}{.913} & \textcolor[rgb]{ .439,  .678,  .278}{.890} & \textcolor[rgb]{ .439,  .678,  .278}{.941} & .910              & \textcolor[rgb]{ .357,  .608,  .835}{.041} & \textcolor[rgb]{ 1,  0,  0}{.884} & \textcolor[rgb]{ 1,  0,  0}{.843} & \textcolor[rgb]{ .357,  .608,  .835}{.815} & \textcolor[rgb]{ 1,  0,  0}{.893} & \textcolor[rgb]{ .357,  .608,  .835}{.856} & \textcolor[rgb]{ 1,  0,  0}{.067} \\
    \hline
    \end{tabular}%
    }
  \label{SOTA}%
\end{table*}%
\begin{figure*}[htb]% The dilation operation bring the influence of weight of edge pixels to the surrounding pixels.
\centering
\includegraphics[width=2\columnwidth]{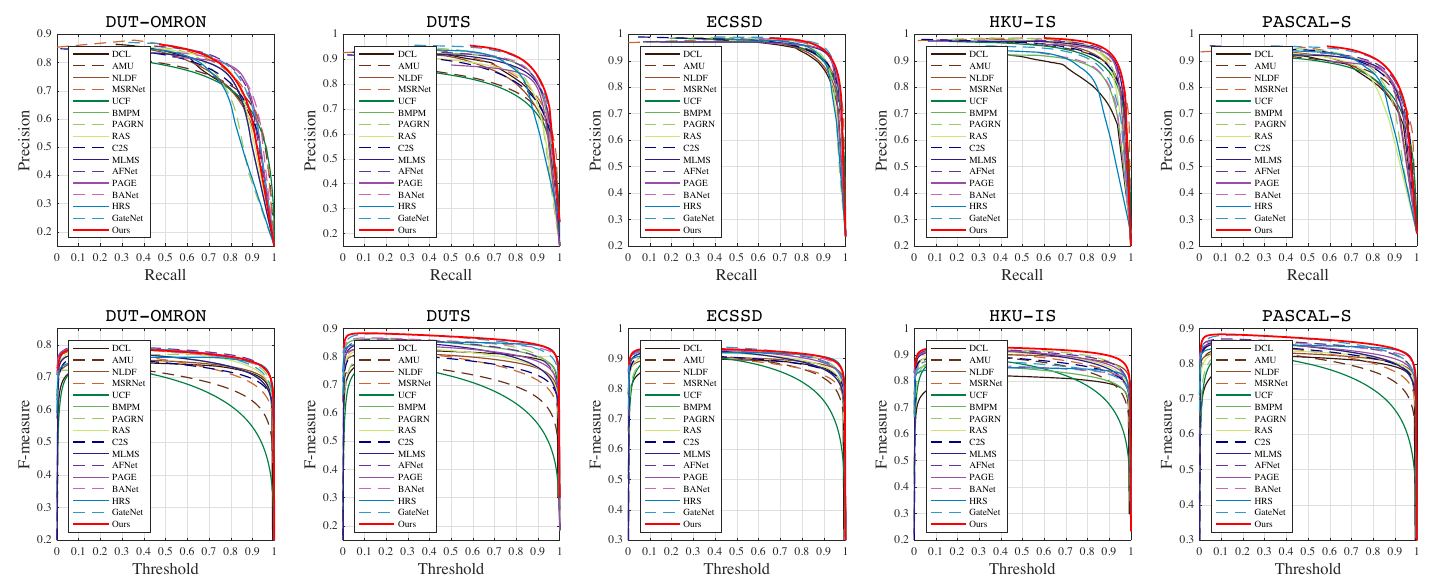} % Reduce the figure size so that it is slightly narrower than the column. Don't use precise values for figure width.This setup will avoid overfull boxes.
\caption{Precision-Recall curves (1st row) and F-measure curves (2nd row) on five common saliency datasets.
}% Here $\lambda$ is 0.
\label{curve}
\end{figure*}
1,4,5 show that it is effective to suppress low-level features with fixed weights, but it is better to regulate the weights according to the analysis results of the PR block.

\subsection{Comparison with State-of-the-arts}
We compare PRNet against 22 SOD state-of-the-art methods, including
DCL~\cite{2016-CVPR-DCL},
%Amulet~\cite{Zhang_2017_ICCV},
NLDF~\cite{Luo2017Non},
%UCF~\cite{2017-CVPR-UCF},
MSRNet~\cite{2017-CVPR-MSRNet},
DSS~\cite{2019-TPAMI-DSS},
BMPM~\cite{2018-CVPR-BMPM},
RAS~\cite{2020-TIP-RAS},
PAGRN~\cite{2018-CVPR-PAGRN},
C2S~\cite{2018-ECCV-C2S},
%MLMSNet~\cite{wu2019a},
PAGE~\cite{2019-CVPR-PAGE},
BANet~\cite{2019-ICCV-BANet},
%HRS~\cite{2019-ICCV-HRS},
AFNet~\cite{2019-CVPR-AFNet}, GateNet~\cite{2020-ECCV-GateNet}, ITSD~\cite{2020-CVPR-ITSD}, FCNet~\cite{2020-NIPS-FCNet}, HVPNet~\cite{2020-TCYB-HVPNet}, CAGNet~\cite{2020-PR-CAGNet}, SAMNet~\cite{2021-TIP-SAMNet}, etc.

For fair comparisons, we use all saliency maps provided by the authors or generated by their codes.
PoolNet~\cite{2019-CVPR-PoolNet} (with StdEdge) adds another dataset (BSDS500) for joint training, which makes the comparison results unfair.
So we compare PRNet with PoolNet (with SalEdge, only DUTS-TE dataset) in Tab.~\ref{SOTA}, and the experimental result shows that our algorithm is better.
Our PRNet (only 130M) is a simple network, so there is no comparison with EGNet (434M)~\cite{2019-ICCV-EGNet} and MINet (650M)~\cite{2020-CVPR-MINet} with large parameters.

\textbf{Quantitative evaluation.} Tab.~\ref{SOTA} shows the scores of the proposed model and 22 state-of-the-art saliency detection methods on five widely used datasets and also demonstrates that the perception-and-regulation strategy is successful in making simple networks perform favorably against other algorithms.
Moreover, the PR curves by our approach outperform other methods, as shown in Fig.~\ref{curve}.

\begin{figure*}[t]% The dilation operation bring the influence of weight of edge pixels to the surrounding pixels.
\centering
\includegraphics[width=1.9\columnwidth,height=12.0cm]{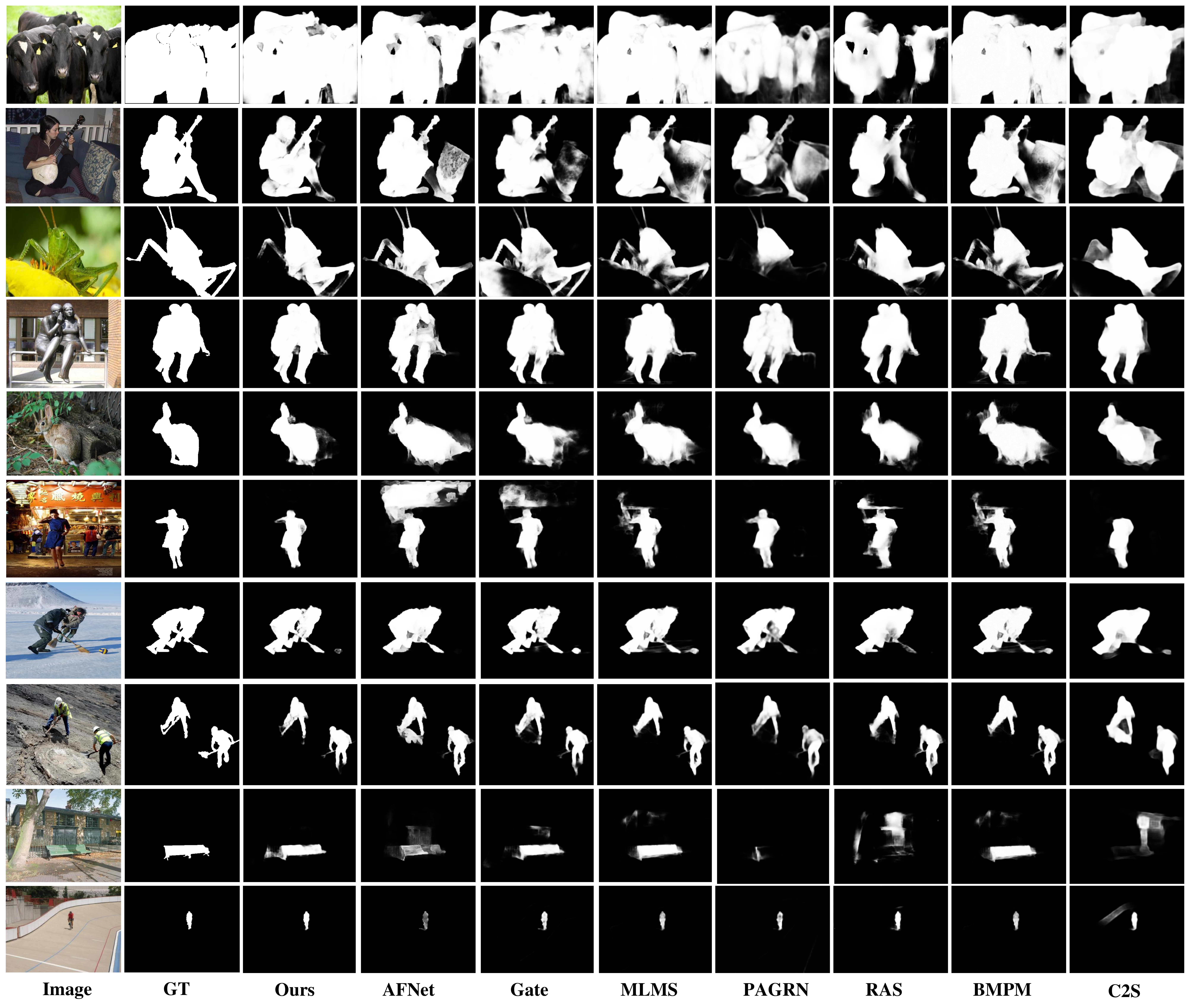} % Reduce the figure size so that it is slightly narrower than the column. Don't use precise values for figure width.This setup will avoid overfull boxes.
\caption{Qualitative comparisons with state-of-the-art algorithms.
From top to bottom, the size of salient objects gradually decreases.
}% Here $\lambda$ is 0.
\label{compare}
\end{figure*}

\begin{figure*}[t]% The dilation operation bring the influence of weight of edge pixels to the surrounding pixels.
\centering
\includegraphics[width=1.9\columnwidth,height=4.0cm]{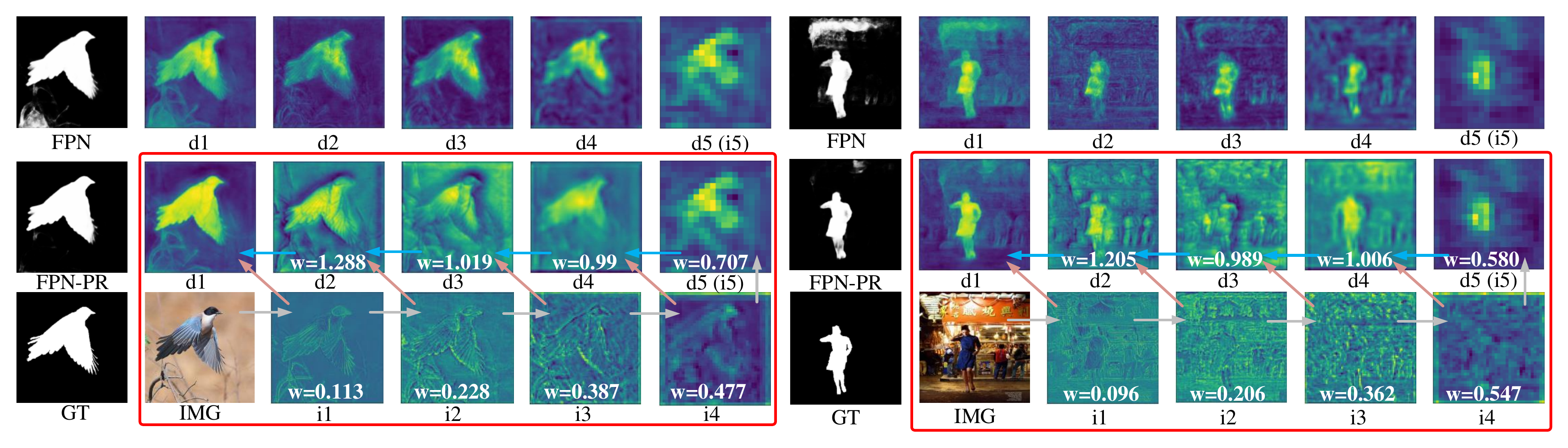} % Reduce the figure size so that it is slightly narrower than the column. Don't use precise values for figure width.This setup will avoid overfull boxes.
\caption{Visual analysis of PR block regulation process.
We show two examples (more examples can be found in the supporting document).
For the feature analysis of the bird,
the 1st and 2nd rows are the decoder features d1-d5 of FPN and the decoder features d1-d5 of FPN-PR$_{FC}$, respectively.
The 3rd row is ground truth, the input image and encoder (VGG-16) features i1-i4.
The 1st column is the final output saliency maps of FPN and FPN-PR$_{FC}$.
In the red box is the feature fusion process of FPN-PR$_{FC}$.
The gray (encoder), red (interlayer features), and blue (decoder features) arrows correspond to the FPN-PR$_{FC}$ in the upper right corner of Fig.\ref{pipline}.
The white font is the result of weight regulated by PR block (PR$_{FC}$).
}% Here $\lambda$ is 0.
\label{tosee}
\end{figure*}

\textbf{Qualitative evaluation.} Fig.~\ref{compare} shows the visual examples produced by our model and other models.
From the 1st row to the 10th row, the size of the salient object gradually changes from the largest to the smallest.
Our algorithm is effective in dealing with multi-scale object detection.
The proposed method performs better in various challenging scenarios, including the small object, medium-sized object, and big object.
Fig.\ref{tosee} shows the visualization results of the whole process.
The difference between the FPN network with a PR block and the FPN network without a PR block is clearly shown.
Besides, we provide some failure cases of our algorithm in the supporting document to help future researchers to conduct further analysis.

% => DEL4
%\begin{figure}[b]
%\centering
%\includegraphics[width=0.9\columnwidth]{img/failurev2.pdf}
%\caption{Failure examples.}
%\label{failure}
%\end{figure}

%\subsection{Failure Case}
%\textcolor{blue}{For future researchers to develop better algorithms, we show some failure cases.
%In the first image of Fig.\ref{failure}, there are many salient objects, and the soldier on the left is not fully captured.
%In the second image, the two sides of the salient object are also not accurately segmented.
%In the third image, the person in the center is gathered with the people around him, which leads to the network not being able to accurately locate the person's body.
%The two recent algorithms CAGNet~\cite{2020-PR-CAGNet} and ITSD~\cite{2020-CVPR-ITSD} do not complete the detection task well.}

% \textcolor{blue}{Because PRNet is a simple network, even if it uses the IEO module to optimize the highest-level feature, there is still room for further optimization.
% The cognitive ability of instance objects may help the network segment salient objects more completely.}

% \textcolor{blue}{In Fig.\ref{failure}, to better analyze the cause of the problem, we visualize the highest-level feature d5 in the PRNet (Fig.\ref{pipline}).
%We find that the positioning result of d5 plays a decisive role in the final output.
%This is because we use the global guidance structure (GGS).
%Besides, in Fig.\ref{FC-S}, we can find that higher-level features are given greater weight.
%  Therefore, improving the accuracy of the high-level features will be an important way to further improve the network performance.}

\section{Conclusions}
In this paper, we propose a novel framework  \textit{PRNet} for salient object detection.
A PR block is designed to help the network understand the global information and assign the feature weights spontaneously and adaptively.
To better perceive semantic information and reasonably allocate weight, we propose 3 perception strategies and carry out comparative experiments.
Through experiments, we verify the different application scenarios of different strategies.
Considering the relationship between local perception and global perception, we propose an IEO module to help the network have the ability to organize a wide space scene and scrutinize highly detailed objects.
% Sufficient experiments conducted on SOD datasets demonstrate that the proposed method performs favorably against 22 state-of-the-art methods.
Sufficient experiments demonstrate that PRNet performs well.
In the future, we may expand PRNet to more complex structures, such as recurrent structure networks and multi-modal SOD networks (RGB-T, RGB-D).
% salient object detection networks (RGB-T, RGB-D).
% In the future, we may further explore the idea of perception-and-regulation block in other network structure, such as multi-modal segmentation network (RGB-T, RGB-D) and recurrent structure network.

\bibliographystyle{IEEEtran}
\bibliography{refer}

\end{document}